\documentclass[10pt,twocolumn,letterpaper]{article}

\usepackage{iccv}              

\definecolor{iccvblue}{rgb}{0.21,0.49,0.74}
\usepackage[pagebackref,breaklinks,colorlinks,allcolors=iccvblue]{hyperref}

\usepackage{multirow}
\usepackage{multicol}
\usepackage{stfloats}
\usepackage{array}

\def\paperID{10218} 
\def\confName{ICCV}
\def\confYear{2025}
\usepackage{colortbl}  
\usepackage{xcolor}
\usepackage{bm}
\title{Learning Robust Stereo Matching in the Wild with Selective Mixture-of-Experts}
\author{
Yun Wang\textsuperscript{1}, Longguang Wang\textsuperscript{2}, Chenghao Zhang\textsuperscript{3}, Yongjian Zhang\textsuperscript{2},\\
Zhanjie Zhang\textsuperscript{4}, Ao Ma\textsuperscript{5}, Chenyou Fan\textsuperscript{6},
Tin Lun Lam\textsuperscript{7,\ddag}, Junjie Hu\textsuperscript{7,\ddag} \\
\textsuperscript{1}City University of Hong Kong, \textsuperscript{2}Shenzhen Campus, Sun Yat-sen University, \\\textsuperscript{3}Chinese Academy of Sciences,
\textsuperscript{4}Zhejiang University, \textsuperscript{5}JD.com, \\\textsuperscript{6}South China Normal University,
\textsuperscript{7}The Chinese University of Hong Kong, Shenzhen\\
{\tt\small ywang3875-c@my.cityu.edu.hk, wanglg9@mail.sysu.edu.cn, zhangchenghao18@mails.ucas.ac.cn,}\\
{\tt\small zhangyj85@mail2.sysu.edu.cn, cszzj@zju.edu.cn, maaoaoma@126.com, }\\
{\tt\small fanchenyou@gmail.com, \{tllam,hujunjie\}@cuhk.edu.cn}
}

\begin{document}
\maketitle

\begin{abstract}
Recently, learning-based stereo matching networks have advanced significantly.
However, they often lack robustness and struggle to achieve impressive cross-domain performance due to domain shifts and imbalanced disparity distributions among diverse datasets.
Leveraging Vision Foundation Models (VFMs) can intuitively enhance the model's robustness, but integrating such a model into stereo matching cost-effectively to fully realize their robustness remains a key challenge.
To address this, we propose SMoEStereo,  a novel framework that adapts VFMs for stereo matching through a tailored, scene-specific fusion of Low-Rank Adaptation (LoRA) and Mixture-of-Experts (MoE) modules.
SMoEStereo introduces MoE-LoRA with adaptive ranks and MoE-Adapter with adaptive kernel sizes. The former dynamically selects optimal experts within MoE to adapt varying scenes across domains, while the latter injects inductive bias into frozen VFMs to improve geometric feature extraction.
Importantly, to mitigate computational overhead, we further propose a lightweight decision network that selectively activates MoE modules based on input complexity, balancing efficiency with accuracy. 
Extensive experiments demonstrate that our method exhibits state-of-the-art cross-domain and joint generalization across multiple benchmarks without dataset-specific adaptation.
The code is available at \textcolor{red}{https://github.com/cocowy1/SMoE-Stereo}.

\end{abstract}    
\section{Introduction}
\label{sec1}

Stereo matching is a critical vision task focused on identifying pixel-wise correspondences between rectified stereo images. It has significant applications in autonomous driving~\cite{chen2015deepdriving}, robot navigation~\cite{murray2000using}, and augmented reality~\cite{wang2023practical}. 
While recent learning-based stereo matching methods have shown impressive performance on standard benchmarks, their generalizability across diverse datasets remains limited.
This is primarily due to significant scene differences and unbalanced disparity distributions across diverse datasets in the wild, potentially leading to noisy and distorted feature maps~\cite{zhang2020domain,jing2023uncertainty}, compromising the robustness of current stereo matching models.

To enhance the robustness of feature representations against domain shifts in stereo matching, we aim to leverage the recent advancements in Vision Foundation Models (VFMs). 
These models, such as DepthAnythingV2~\cite{yang2024depth_v2} for monocular depth estimation and SegmentAnything~\cite{kirillov2023segment} for segmentation, are built on Vision Transformers (ViTs) pre-trained on large-scale diverse datasets. 
Such VFMs have demonstrated great effectiveness in providing robust, general-purpose deep features in many vision tasks.
However, fully unleashing their potential for robust cross-domain stereo matching remains limited, and thus two intuitive limitations emerged:
First, directly applying VFMs for stereo matching exhibits limited zero-shot performance. 
Although VFMs are highly effective at extracting semantic information from tasks such as single-image segmentation/classification/regression, they struggle to generate discriminative features necessary for precise similarity measurements in dense cross-view feature matching~\cite{weinzaepfel2023croco,liu2024playing,zhanglearning2024}, as illustrated in Fig.~\ref{sec1:teaser} (c). 
\textcolor{black}{Second, existing fine-tuning methods such as Low-Rank Adaptation~\cite{hu2022lora} or small decoder~\cite{zhanglearning2024} struggle to address the varying complexity of real-world in-the-wild stereo scenes, as they tend to employ a uniform low-rank subspace or a fixed CNN decoder across all inputs, treating vastly heterogeneous domains with rigid, one-size-fits-all feature refinement. This inflexibility limits their capacity to dynamically adjust to scene-specific characteristics, resulting in suboptimal generalization for in-the-wild scenarios~\cite{chen2023benchmarking,shen2021cfnet,Wei_2024_CVPR}.
}

\textcolor{black}{To tackle the above problems, we propose incorporating Low-Rank Adaptation (LoRA) with varying ranks into a Mixture-of-Experts (MoE) design to adapt VFMs for robust stereo matching with minimal costs.
Rather than naively using LoRA with a fixed rank, which limits its adaptability to diverse scenes for robust stereo matching,
we extend conventional LoRA by developing a scene-conditioned selection mechanism that dynamically selects optimal low-rank subspaces from a spectrum of predefined 
rank values, enabling adaptive feature refinement tailored to in-the-wild scene characteristics.}
\textcolor{black}{We also observe that frozen VFMs, even with learnable LoRA layers, lack intrinsic Inductive Bias, essential for modeling local visual structures for stereo matching~\cite{li2021revisiting, guo2022context}.
To bridge this gap, we introduce Inductive Bias by integrating Convolutional Neural Networks (CNNs) into each ViT block. 
Similarly, we embed MoE modules with several CNN adapters of varying receptive fields in each ViT block to incorporate Inductive Bias, while enhancing the model’s ability to capture local geometric structures.
This hybrid design achieves complementary feature learning: CNN branches emphasize fine-grained local geometric details, while LoRA pathways model long-range interactions. As a result, it reduces stereo matching D1 error up to 30\% versus vanilla VFM-LoRA baselines, significantly improving model robustness.}
\textcolor{black}{Besides, integrating MoE into all ViT blocks introduces redundancy and extra costs, posing a critical bottleneck for deployment efficiency, a paramount concern in stereo matching applications.} To this end, we introduce a lightweight decision network into each MoE layer. This network predicts binary decisions to activate MoE modules, saving computational costs by discarding redundant modules for simple samples and using more modules for complex ones. The decision network is jointly optimized with our MoE modules, incorporating a usage loss to measure computational costs and encourage policies that reduce redundancy while maintaining accuracy. 
\textcolor{black}{The hyper-parameter $\gamma\in(0,1]$ can regulate the overall computational budget by scaling the activated MoE modules' computational load proportionally, enabling flexible adaptation to varying resource constraints,  critical for mobile devices with diverse computation demands~\cite{wang2019anytime}.}

Since both MoE modules and the experts within each MoE can be \textit{selectively} activated to adapt to different input characteristics, we refer to our method as SMoEStereo.
This method fully unleashes the potential of existing VFMs for stereo matching, as illustrated in Fig.~\ref{sec1:teaser}. 
Extensive experiments show that SMoEStereo exhibits strong robustness, achieving state-of-the-art cross-domain generalization performance on the KITTI, Middlebury, ETH3D, and DrivingStereo datasets. 
Additionally, it achieves state-of-the-art joint generalization on the ETH3D, KITTI, and Middlebury benchmarks using the same trained model without any adaptation.
The key contributions are as follows:

\begin{itemize}
    \item 
    \textcolor{black}{We present SMoEStereo, an efficient yet powerful approach that leverages pre-trained Vision Foundation Models for robust stereo matching with minimal cost.}
\end{itemize}

\begin{itemize}
    \item \textcolor{black}{We propose integrating MoE LoRA with varying ranks and MoE Adapter layers with varying kernel sizes into VFMs. These tailored designs facilitate scene-specific adaptation, enabling robust stereo matching across diverse real-world scenarios.}

\end{itemize} 

\begin{itemize}
    \item We design a lightweight decision network integrated within each MoE module, which dynamically selects the relevant modules and deactivates less critical ones. This mechanism balances model accuracy and efficiency, enabling flexible adaptation to diverse resource constraints.
\end{itemize}

\begin{itemize}
    \item 
    Our approach exhibits strong cross-domain generalization and excels across multiple benchmarks using the same fixed model without further adaptation, significantly outperforming previous robust counterparts.

\end{itemize}


\section{Related Work}
\label{sec2}

\subsection{Robust Stereo Matching}
Recently, following the success of RAFT~\cite{teed2020raft} in optical flow tasks, iterative approaches such as RAFTStereo~\cite{Lipson2021RAFTStereoMR}, IGEVStereo~\cite{li2022practical}, and Selective-IGEV~\cite{wang2024selective} have set new benchmarks by iteratively updating the disparity field through correlation volume sampling.
Despite advancements in model design, achieving robust performance across varied scenarios remains challenging.


To tackle this, robust stereo matching methods have been increasingly studied, broadly divided into two categories:
1) Cross-domain Generalization: This category focuses on the network's generalization to unseen scenes. Previous works try to tackle the problem by introducing domain normalization~\cite{zhang2019ga}, leveraging pre-trained features on ImageNet~\cite{liu2022graftnet}, or developing diverse training strategies to learn domain-invariant features~\cite{chang2023domain,chuah2022itsa,zhang2022revisiting}.
2) Joint Generalization: This category aims to push the network to perform consistently well on various datasets without retraining. 
CFNet~\cite{shen2021cfnet} and its improved UCFNet~\cite{2023uCFNet} introduce a cascaded and fused cost volume-based network to handle domain differences.
CREStereo++~\cite{jing2023uncertainty} introduces an uncertainty-guided adaptive warping module to enhance the robustness of the recurrent network in different scenarios. 
LoS~\cite{likh2024los} integrates structure information to improve model performance in challenging areas.
However, these robust methods typically rely on classic feature extractor backbones such as ResNet~\cite{he2016deep}, UNet~\cite{ronneberger2015u}, and Feature Pyramid Network~\cite{lin2017feature}, which suffer from limited receptive fields. Consequently, the potential of more powerful Vision Foundation Models (VFMs) based on Vision Transformers (ViTs) in robust stereo matching remains unexplored.

\subsection{Parameter-Efficient Fine-tuning (PEFT)}
Visual Foundation Models (VFMs) are defined as base vision models trained on large-scale data through self/semi-supervised learning, designed for adaptation to downstream vision tasks~\cite{awais2023foundational}.
Recently, they~\cite{rombach2022high,dust3r_cvpr24,kirillov2023segment,yang2024depth_v2,oquab2024dinov2} have emerged as a solution to enhance model discriminability and robustness.
Given the substantial computational costs of fully fine-tuning VFMs, Parameter Efficient Fine-Tuning (PEFT) has emerged as a promising alternative. Notable examples include Visual Prompt Tuning (VPT)~\cite{jia2022visual}, which augments inputs with extra learnable tokens; AdaptFormer~\cite{chen2022adaptformer}, proposing AdaptMLP to replace the MLP block; Adapter-tuning~\cite{liu2024playing,zhanglearning2024}, adding lightweight decoder modules; and Low-Rank Adaptation (LoRA)~\cite{hu2021lora}, injecting trainable rank decomposition matrices into transformer layers. 
Among these, the work of Zhang et al.~\cite{zhanglearning2024} is most closely related to ours.
They adapt VFMs by proposing a feature adapter to obtain robust features. 
However, our SMoE differs in two key aspects: (1) Instead of a fixed CNN decoder, SMoE uses rank-adaptive LoRA and kernel-adaptive CNNs to encode local and global cues, enabling scene-specific adaptation; (2) Unlike their computationally rigid pipeline requiring sequential processing, SMoE dynamically skips non-critical MoE modules for simple scenes, boosting efficiency and real-world applicability.


\subsection{Mixture of Experts (MoE)}
\textcolor{black}{Mixture-of-Experts (MoE)~\cite{jacobs1991adaptive, shazeer2017outrageously} is designed to expand model capacity while introducing small computational overhead.
Sparse MoE~\cite{shazeer2017outrageously} introduces a router to select a subset of experts, using the gating network to regulate sparsity for computational savings.
Feed Forward Networks (FFN) are commonly employed as the default choice for experts~\cite{shazeer2017outrageously, riquelme2021scaling,zhou2022mixture, bao2022vlmo,fedus2022switch}.
Recently, several studies combine MoE with multiple uniform LoRAs into large language/vision models by developing diverse routing mechanisms~\cite{luo2024moelora,li2024mixlora,dou2024loramoe,liu2024adamole,yang2024multi}, enabling multi-task learning.}

\textcolor{black}{While previous works on MoE aim to expand model capacity or enable multi-task learning,
our SMoE focuses on dynamically selecting optimal experts to adapt to diverse in-the-wild scenes, thereby enhancing robust stereo matching.
Overall, our work differs from prior studies in three aspects:
1) Unlike homogeneous MoE experts, our SMoE employs heterogeneous experts with varying-rank LoRA or varying-kernel CNN configurations to handle diverse in-the-wild scenarios.
2) While MoE is mostly employed during pre-training, we integrate it as a parameter-efficient tuning mechanism for stereo matching task.
3) We novelly introduce a decision network to selectively activate the most suitable MoE modules, enhancing efficiency and applicability for stereo matching.}


\section{Methodology}

    \begin{figure*}[t]
    \centering
    \includegraphics[width=1\linewidth]{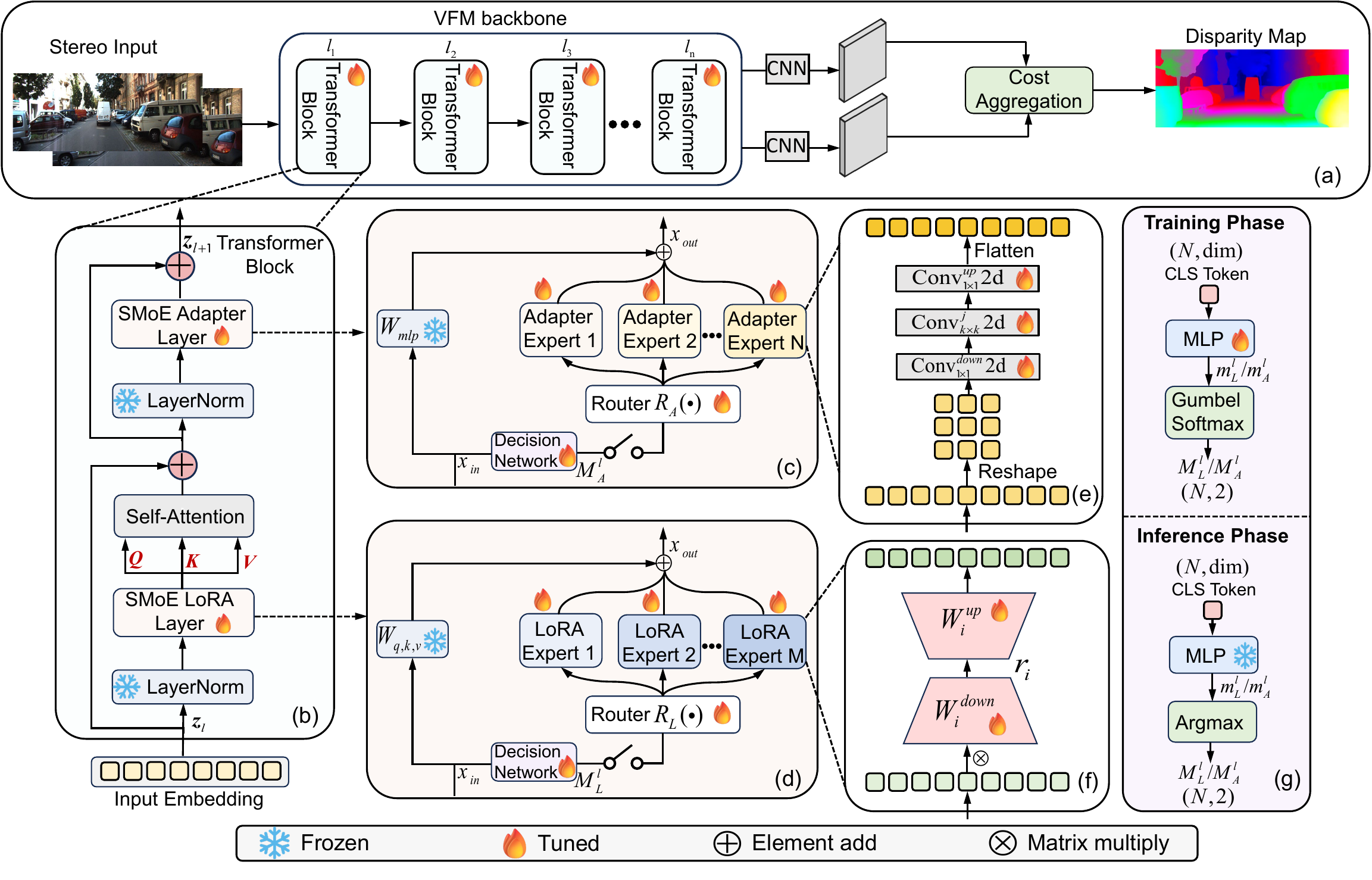}
    \vspace{-0.7cm}
    \caption{Overview of the designed SMoEStereo framework. (a) Overall model structure; (b) Overall the transformer block; (c) Details of the designed SMoE Adapter layer; (d) Details of the designed SMoE LoRA layer; (e) Details of each Adapter expert; (f) Details of each LoRA expert; (g) Details of the decision network. Note that our SMoE can serve as a plug-and-play module for most stereo networks.} 
    \label{sec3:framework}
    \vspace{-0.2cm}
\end{figure*}

\subsection{Overview}
RAFT-Stereo~\cite{Lipson2021RAFTStereoMR} is used as our backbone.
We replace its feature extractor with VFMs while the remaining structures are unchanged.
Specifically, the network first extracts stereo features using the well-designed VFM from stereo pairs. We modify the ViT blocks by integrating the designed MoE LoRA and Adapter layer with the vanilla blocks. {A shallow CNN block~\footnote{The shallow CNN block comprises several CNN residual layers.}  is then used to compress the feature dimension and enhance the locality.}
Then, a correlation volume pyramid is generated by computing the inner product along the epipolar lines.
A multi-level GRU network is subsequently introduced to perform cost aggregation by recurrently updating the disparity field from the correlation volume pyramid.  
A sketch of the proposed pipeline is shown in Fig.~\ref{sec3:framework}.

\subsection{Selective Mixture of Expert (SMoE)}
As described in Sec.~\ref{sec1}, the proposed Selective MoE integrates MoE LoRA and MoE Adapter layers within each ViT block.
Besides, a layer decision network is inserted before each of the MoE layers, and it is trained to produce reasonable usage policies for MoE layer selection.
In this way, our SMoEStereo learns to adaptively choose 1) which LoRA and Adapter experts to use and 2) which MoE layer to skip based on per input to improve the model efficiency.

\noindent \textbf{MoE LoRA Layer.}
Vanilla LoRA~\cite{hu2021lora} typically pre-define a matrix rank to update the learnable weights $\Delta W$ while freezing the pre-trained weights $W$.
However, in practical deployment environments, it is challenging to pre-define an ideal rank for different test samples.
To address this, we introduce a MoE LoRA layer architecture that employs various LoRA layers with different matrix ranks as experts, thereby identifying the optimal LoRA expert for each input query.
As shown in Fig.~\ref{sec3:framework} (d), the MoE LoRA layer contains $M$ experts, denoted as $\left\{E^{i}_{L}\right\}\sum^{M}_{i=1}$. Each LoRA expert ${E^{i}_{L}}(\cdot)$ corresponds to a specific rank $r_{i}$. We then design a router network  $R_{L}(\cdot)$ to dynamically select the Top-\textit{k} optimal experts (default: $k$=1).
Specifically, given input $x_{in}\in\mathbb{R}^{N\times dim}$ is fed into the MoE LoRA layer, where $N$ is the token count and $dim$ is the feature dimension, the forward propagation of the MoE-LoRA layer is defined as:
\vspace{-0.2cm}
\begin{equation}
x_{out} =  W_{q,k,v}\; x_{in}  + \sum_{i=1}^{M} R_{L}(x_{in})\cdot E^{i}_{L}(x_{in}),
\label{eq:moe_lora}
\end{equation}
where the pre-trained weights $W_{q,k,v}$ is frozen during training and $W_{q,k,v}\in\left\{W_{q}, W_{k}, W_{v}\right\}$ could be either query, key, or value matrices. 
Each LoRA expert is formulated by:
\begin{equation}
   E^{i}_{L}(x_{in}) =\Delta W  x_{in} = W^{up}_{i}W^{down}_{i} x_{in},
   \label{lora}
\end{equation}
where $W^{up}_{i}\in\mathbb{R}^{d\times dim}$ and  $W^{down}_{i}\in\mathbb{R}^{r_{i}\times r_{i}}$ are two trainable
matrices, and $r_{i} \ll \left\{D, dim\right\}$.
The router mechanism is presented as follows: 
\vspace{-0.2cm}
\begin{equation}
R(x_{in}) ={\rm Topk}(\frac{\exp (W^{router} x_{in}/ \tau)}{\sum_{k=1}^M \exp (W^{router} x_{in}/ \tau)}, k),
\label{gate}
\vspace{-0.1cm}
\end{equation}
where $W^{router}$ represents the slim trainable weights, and $\tau$ denotes the temperature, empirically set to 5.
In addition to the MoE LoRA layer, a router mechanism is also employed for the subsequent MoE Adapter layer.

\noindent \textbf{MoE Adapter Layer.}
Existing ViTs struggle to learn from small datasets due to a lack of intrinsic inductive bias in modeling local visual structures~\cite{raghu2021vision, park2022vision, zhang2025spast,zhang2025lgast,xu2021vitae}.
To address this, we propose an MoE Adapter layer that captures local spatial priors with different receptive fields before injecting them into the plain ViT blocks.
As shown in Fig.~\ref{sec3:framework} (c), the designed MoE Adapter layer consists of several CNN Adapter experts $\left\{E^{j}_{A}\right\}\sum^{N}_{j=1}$ with different kernel sizes $k$ to embed local geometry into tokens. Similar to the router mechanism in the MoE LoRA layer, we introduce a router network $R_{A}(\cdot)$ to select the optimal experts dynamically.
The output can be formularized as:
\vspace{-0.1cm}
\begin{equation}
x_{out} =  W_{mlp}\; x_{in} + \sum_{j=1}^{N} R_{A}(x_{in})\cdot E^{j}_{A}(x_{in}),
\label{eq:moe_adapter}
\vspace{-0.1cm}
\end{equation}
where $W_{mlp}$ is frozen as the pre-trained weights during training. For each CNN Adapter expert $E_{A}(\cdot)$, it can be calculated as (we omit the
activation layer):
\begin{equation}
E_{A}^{j}(x_{in})={\rm Conv}_{1 \times 1}^{up}({\rm Conv}_{k \times k}^{{j}}({\rm Conv}_{1 \times 1}^{down}(x_{in}))), 
\label{adapter}
\end{equation}
where ${\rm Conv}^{down}_{1\times 1}$ and ${\rm Conv}^{up}_{1\times 1}$
are two 1 × 1 convolution layers, which reduce and restore the number of channels, respectively.
${\rm Conv}^{j}_{k\times k}$ indicates the specific convolution layer with kernel size $k$ in
different experts.

Meanwhile, to prevent the gating network from assigning large weights to the same few experts, we apply a soft constraint~\cite{shazeer2017outrageously} on the batch-wise average of each MoE LoRA and MoE Adapter router $R(\cdot)$.
As such, for a given batch of data $X$, the MoE balance loss $\mathcal{L}_{blc}$ is defined as:
\vspace{-0.05cm}
\begin{equation}
\mathcal{L}_{blc}  = \frac{\sigma^{2}({Q(X)})}{\mu(Q(X))}; \qquad
Q(X) = \sum_{x \in X} R(x) ,
\vspace{-0.05cm}
\end{equation}
where $\sigma^{2}(Q(x))$ and $\mu(Q(X))$ represent the variance and mean of $Q(x)$, respectively, and we use $Q (x)$ to define the importance of experts with the MoE layer.
Consequently, the balance loss $\mathcal{L}_{blc}$ ensures equal importance among all experts, fostering their diverse utilization.

\begin{table}[t]
\centering
\scriptsize
\caption{Domain generalization evaluation on real-world datasets. $^\ddag$ denotes that VFMs use the ViT-large model capacity. (The best results in \textbf{bold} and the sub-optimal best results in \textbf{\textcolor{blue}{blue}}).}
\vspace{-0.3cm}
\setlength{\tabcolsep}{1.5mm}{
    \begin{tabular}{c|c|c|c|c}
    \toprule
    \multirow{2}{*}[-1pt]{Method} &  \multirow{1}{*}[-1pt]{\textbf{KIT 2012}}  &  \multirow{1}{*}[-1pt]{\textbf{KIT 2015}}  &  \multirow{1}{*}[-1pt]{\textbf{Middle}} &  \multirow{1}{*}[-1pt]{\textbf{ETH3D}}\\
    & \textbf{Bad 3.0} & \textbf{Bad 3.0} & \textbf{Bad 2.0}  & \textbf{Bad 1.0}\\
    \midrule
    CFNet~\cite{shen2021cfnet}  & 4.7 & 5.8 & 15.3 & 5.8 \\
    RAFT-Stereo~\cite{Lipson2021RAFTStereoMR}& 5.1 & 5.7 & 12.6 & ${3.3}$ \\ 
    UCFNet\_pretrain~\cite{2023uCFNet}  & 4.5 & 5.2 & 26.0 & 4.8 \\
    CREStereo++~\cite{jing2023uncertainty} & ${4.7}$ & 5.2 & 14.8 & ${4.4}$ \\
    DLNR~\cite{zhao2023high} & 9.1 & 16.1 & 9.8& 23.0\\
    LoS~\cite{li2024local} & 4.4 & 5.5 & 19.6 & {3.1} \\
    Selective-IGEV~\cite{wang2024selective} & 5.7 & 6.1 & 13.3 & 6.1\\
    MochaStereo~\cite{chen2024mocha} & 4.9 & 5.9 & 11.5 & 3.9 \\
    Former-PSMNet$^\ddag$ (SAM)~\cite{zhanglearning2024}& 4.3& 5.0 &9.5 & 6.4 \\
    Former-CFNet$^\ddag$ (DINOv2)~\cite{zhanglearning2024} &4.0 & 5.0 & 8.8 & 4.5 \\
    Former-RAFT$^\ddag$ (DAM) ~\cite{zhanglearning2024} & \textbf{3.9} & {5.1} & {8.1} & 3.3\\
    \midrule
    \textbf{SMoEStereo (DINOV2~\cite{oquab2024dinov2})} & 4.39 & 4.97 & 7.56 & {2.57}\\
    \textbf{SMoEStereo (SAM~\cite{kirillov2023segment})} & 4.27 & \textbf{\textcolor{blue}{4.89}} & \textbf{\textcolor{blue}{7.10}} &  \textbf{2.07}\\
    \textbf{SMoEStereo (DAM~\cite{depthanything})} & 4.44 & 5.05 & 7.57 & 2.56\\
    \textbf{SMoEStereo (DAMV2~\cite{yang2024depth_v2})} & $\textbf{\textcolor{blue}{4.22}}$ & $\textbf{4.86}$ & \textbf{7.05} & \textbf{\textcolor{blue}{2.10}}\\
    \midrule
    \end{tabular}}
\label{sec4:cross_comparison}
\vspace{-0.5cm}
\end{table}

\noindent \textbf{Decision Network.}
Previous studies~\cite{lee2022surgical,devoto2024adaptive} demonstrated that different layers across the ViTs have varying contributions to the overall performance, depending on the input data distributions. 
Therefore, the goal of the decision policy is to identify `important' MoE layers within each ViT block (i.e., 0 for `unimportant' modules and 1 for `important' ones). To achieve this, a decision network with few learnable parameters is introduced to reduce redundancy by selectively keeping or dropping MoE layers based on the input samples.
Specifically, the class token input $x_{cls}\in \mathbb{R}^{N\times dim}$ is first fed to MLP to produce the corresponding probability vectors $m^{l}_{L}$ for MoE LoRA layer or $m^{l}_{A}$ for MoE Adapter Layer, respectively.
Then, to make the whole framework end-to-end trainable, the Gumbel softmax trick~\cite{maddison2016concrete} is used to obtain the softened binary masks at \textit{l}-th layer as follows:
\begin{equation}
\mathbf{\textit{M}}^{l}_{i}=\frac{\exp \left(\ \left(\mathbf{\textit{m}}^{l}_{i}+G_{i}\right) / \tau\right)}{\sum_{j=1}^K \exp \left(\ \left(\mathbf{\textit{m}}^{l}_{j}+G_{j}\right) / \tau\right)} \\
\text { for } i=1,...K ,
\end{equation}
where $K$ represents the total number of categories ($K = 2$ for binary decisions in our case), $G_{i}$ is a Gumbel noise tensor with all elements sampled from $\mathrm{Gumbel (0, 1)}$, and $\tau$ controls the smoothness of $M_{i}^l$. During training, we use the first element, $M_{1}^l$, while binary values are obtained via \textit{Argmax} during inference, as shown in Fig.~\ref{sec3:framework} (g). 
For simplicity, we denote the usage policies for the MoE LoRA and MoE Adapter layers as $M^{l}_{L}$ and $M^{l}_{A}$, respectively.
Thus, Eq.~\eqref{eq:moe_lora} \&~\eqref{eq:moe_adapter} at the $l$-th layer can be reformulated as:
\begin{equation}
\vspace{-0.05cm}
\setlength\abovedisplayskip{3pt}
\setlength\belowdisplayskip{3pt}
\begin{aligned}
x_{out} &= W_{q,k,v} x_{in}   +  M^{l}_{L} (\sum_{i=1}^{M} R_{L}(x_{in})\cdot E^{i}_{L}(x_{in})) \\
x_{out} &= W_{mlp} x_{in} + M^{l}_{A} (\sum_{j=1}^{N} R_{A}(x_{in})\cdot E^{j}_{A}(x_{in})) ,
\end{aligned}
\vspace{-0.05cm}
\end{equation}

\begin{table}[t]
\centering
\scriptsize
\caption{Zero-shot performance on DrivingStereo under different weathers, evaluated with official weights and the D1 metric.}
\vspace{-0.3cm}
\setlength{\tabcolsep}{2.5mm}{
    \begin{tabular}{c|c|c|c|c|c}
    \toprule
    {Method} &  \multirow{1}{*}[-1pt]{\textbf{Sunny}}  &  \multirow{1}{*}[-1pt]{\textbf{Cloudy}}  &  \multirow{1}{*}[-1pt]{\textbf{Rainy}} &  \multirow{1}{*}[-1pt]{\textbf{Foggy}} & \textbf{Avg.} \\
    \midrule
    CFNet~\cite{shen2021cfnet}  & 5.4 & 5.8 & 12.0 &  6.0 & 7.3 \\
    PCWNet~\cite{shen2022pcw} & 5.6 & 5.9 & 11.8 & 6.2 & 7.4 \\
    DLNR~\cite{zhao2023high} & 27.1 & 28.3 & 34.5 & 29.0 & 29.8 \\
    IGEVStereo~\cite{xu2023iterative} & 5.3 & 6.3 & 21.6 & 8.0 & 10.3 \\
    Selective-IGEV~\cite{wang2024selective} & 7.0 & 8.0 & 18.4 & 12.9 & 11.1\\
    MochaStereo~\cite{chen2024mocha} & 12.8 &27.4 & 24.6 & 22.8 & 21.9 \\
    Former-CFNet$^\ddag$~\cite{zhanglearning2024} &$\textcolor{blue}{\textbf{3.8}}$ & \textbf{2.7} & \textcolor{blue}{\textbf{8.3}} & \textcolor{blue}{\textbf{5.2}} & $\textcolor{blue}{\textbf{5.0}}$\\
    \midrule
    \textbf{SMoEStereo} & $\textbf{3.3}$ & $\textcolor{blue}{\textbf{3.0}}$ & \textbf{6.0} & \textbf{4.7} & \textbf{4.3}\\
    \midrule
    \end{tabular}}
\label{sec4:weather_comparison}
\vspace{-0.45cm}
\end{table}

\begin{table}[t]
\centering
\scriptsize
\caption{Inference efficiency comparison with VFM-based generalized methods. $\dag$ indicates evaluation using the author's codes. $\ast$ denotes extra activated parameters within VFMs during inference.}
\vspace{-0.3cm}
\setlength{\tabcolsep}{0.2mm}{
    \begin{tabular}{c|c|ccc}
    \toprule
    {Method} & Capacity &  {\textbf{Memory (GB)}} & {\textbf{Time (s)}}  &  {\textbf{Params. $^\ast$ (M)}}  \\
    \midrule
     Former-PSMNet (SAM)$^\dag$~\cite{zhanglearning2024} & ViT-Large & 6.0 & 0.52 & 6.9 \\ 
    Former-RAFT (DAM)$^\dag$~\cite{zhanglearning2024} & ViT-Large & 4.1 & 0.47 & 6.9 \\ 
    \midrule
    \textbf{SMoEStereo (SAM)} & ViT-Base & \textcolor{blue}{\textbf{2.0}}  &  \textcolor{blue}{\textbf{0.20}} & \textcolor{blue}{\textbf{4.06}}   \\
    \textbf{SMoEStereo (DAM)} & ViT-Base & \textbf{1.9} & \textbf{0.18} & \textbf{2.86} \\
    \bottomrule
    \end{tabular}}
\label{sec4:vfm_comparison}
\vspace{-0.4cm}
\end{table}
In summary, the decision network generates usage policies for each MoE layer within transformer blocks based on the input. The input is then processed through the block according to these policies.
Meanwhile, to encourage reducing the overall computational cost,
we devise the usage loss as:
\vspace{-0.1cm}
\begin{equation}
\mathcal{L}_{usage} = (\frac{1}{L} \sum_{l=1}^{L}M^{l}_{L}-\gamma)^{2} + (\frac{1}{L} \sum_{l=1}^{L}M^{l}_{A}-\gamma)^{2},
\vspace{-0.1cm}
\end{equation}
here, $L$ denotes the total number of blocks of the  ViT backbone. 
The hyper-parameters $\gamma \in (0, 1]$ indicate target computation budgets regarding the percentage of blocks to keep.

\noindent \textbf{Total Loss Function.}
Following~\cite{Lipson2021RAFTStereoMR}, we supervise the $L1$ distance between the sequence of disparity predictions $\left\{\hat{D}_{1}, ..., \hat{D}_{N}\right\}$ and the Ground Truth Disparity $D_{gt}$:
\vspace{-0.1cm}
\begin{equation}
\mathcal{L}_{disp}=\sum^{N}_{i=1}\beta^{N-i}||(D_{gt}-\hat{D}_{i})||_{1},
\end{equation}
where $N$ = 16 during training and the exponentially weight $\beta$ is set to 0.9. 
The total loss is formulated as follows:
\begin{equation}
\mathcal{L}_{total} = \mathcal{L}_{disp} + \lambda_{1} \mathcal{L}_{blc} + \lambda_{2} \mathcal{L}_{usage}.
\label{eq:total_loss}
\end{equation}
Where we set the hyper-parameters to $\lambda_{1}=\lambda_{2}=1$.

\section{Experiments}
\subsection{ Experimental Settings}

\begin{table}[t]
\centering
\scriptsize
\caption{Performance comparison of the proposed SMoE against other finetuning methods. Params (M) Train/Test refers to the number of learnable and additional activated parameters within the VFM backbone for the training and inference phases, respectively.}
\vspace{-0.2cm}
\setlength{\tabcolsep}{0.2mm}{
    \begin{tabular}{c|c|c|c|c|c|c}
    \toprule
    \multirow{2}{*}[-1pt]{Backbone} & {Fine-tune} & { Params (M)} & \multirow{1}{*}[-1pt]{\textbf{KIT 2012}} & \multirow{1}{*}[-1pt]{\textbf{KIT 2015}}  &  \multirow{1}{*}[-1pt]{\textbf{Middle}} &  \multirow{1}{*}[-1pt]{\textbf{ETH3D}}\\
    & Method & Train/Test  & \textbf{Bad 3.0} & \textbf{Bad 3.0} & \textbf{Bad 2.0}  & \textbf{Bad 1.0}\\
    \midrule
    \multirow{6}{*}[-1pt]{DAMV2~\cite{yang2024depth_v2}} & Frozen & 0/0 & 11.7 & 15.4 & 24.6 & 16.2 \\
     & Full-Finetuning & 97.5/0 & 10.1 & 14.3 & 22.3 & 15.7\\
      & VPT~\cite{jia2022visual} & 4.72/4.72 & 4.65 & 5.18 & 8.44 & 2.71 \\
    & AdapterFormer~\cite{chen2022adaptformer} & 4.14/4.14 & 4.59 & {5.11} & 8.39 & 2.77\\ 
      (ViT-base) & AdapterTuning~\cite{zhanglearning2024} & 4.20/4.20 & \textcolor{blue}{\textbf{4.40}} & 5.09 & 8.03 & \textcolor{blue}{\textbf{2.61}}\\
      & LoRA~\cite{hu2021lora} & 4.72/4.72 & 4.47 & \textcolor{blue}{\textbf{5.03}} & \textcolor{blue}{\textbf{7.67}} & 2.83\\
    & \textbf{SMoE (Ours)} & 6.81/2.86 & $\textbf{4.22}$& $\textbf{4.86}$ & $\textbf{7.05}$ & \textbf{2.10} \\
    \midrule
    \multirow{6}{*}[-1pt]{SAM~\cite{kirillov2023segment}} & Frozen & 0/0 & 10.3 & 14.8 & 22.8 & 16.4 \\
     & Full-Finetuning & 93.7/0 & 8.91 & 10.5 & 18.7 & 13.6 \\
      & VPT~\cite{jia2022visual} & 4.72/4.72  & 4.60  & 5.19 & 8.21 & \textbf{\textcolor{blue}{2.39}}  \\
    & AdapterFormer~\cite{chen2022adaptformer} & 4.14/4.14 & 4.53 &  \textbf{\textcolor{blue}{4.95}} &8.85 & 2.99 \\ 
    (ViT-base) & AdapterTuning~\cite{zhanglearning2024} & 6.91/6.91 & 4.58  & 5.11 & 7.82 & 2.53 \\
      & LoRA~\cite{hu2021lora} & 4.72/4.72 & \textcolor{blue}{\textbf{4.49}} & 5.17 & \textcolor{blue}{\textbf{7.76}} & 2.78\\
    & \textbf{SMoE (Ours)} & 6.81/4.06 & \textbf{4.27} & \textbf{4.89} & \textbf{7.10} & \textbf{2.07} \\
    \midrule
    \end{tabular}}
\label{sec4:peft}
\vspace{-0.5cm}
\end{table}

\noindent \textbf{Datasets \& Metrics.}
SceneFlow~\cite{mayer2016large} is a large-scale synthesis dataset containing 35454 training samples and  4370 validation samples. 
KITTI 2015~\cite{menze2015object} and KITTI 2012~\cite{geiger2012we} are two real-world datasets with 200 and 194 outdoor driving scenes stereo images, respectively. 
Middlebury~\cite{middlebury2014} consists of 28 training samples and 15 high-resolution evaluation images of indoor scenes. 
ETH3D~\cite{Schps2017AMS} is a low-resolution dataset with 27 gray images. 
We also qualitatively evaluate the robustness of our network on the DrivingStereo dataset~\cite{yang2019drivingstereo} under challenging weather conditions. 
Except for Middlebury (half-resolution), we use full resolution for these datasets.
For cross-domain generalization, we use the SceneFlow dataset to pre-train our model and report its cross-domain generalization ability on realistic datasets.
For joint generalization evaluation, we strictly follow previous Robust Vision Challenging (RVC) settings~\cite{shen2021cfnet,2023uCFNet,jing2023uncertainty}, where we adopt KITTI 2015 \& 2012, Middlebury, and ETH3D training sets to finetune our pre-trained model jointly and evaluate on three real-world public benchmarks (KITTI 2015, Middlebury, and ETH3D) using a single fixed model without adaptation.
Following previous works~\cite{wang2024cost,wang2025adstereo,guo2022cvcnet,wang2022spnet}, EPE and $t$ px (the percentage of outliers with an absolute error greater than $t$ pixels) are used to evaluate the performance. 

\noindent \textbf{Implementation Details.}
DAMV2~\cite{yang2024depth_v2} (ViT-base) and RAFT-Stereo~\cite{Lipson2021RAFTStereoMR} are used as final backbones to present SMoEStereo in the experiments. 
We also adopt DAM~\cite{depthanything}, SAM~\cite{kirillov2023segment}, and DINOV2~\cite{oquab2024dinov2} as VFM variants.
During pre-training, we pre-train our model on SceneFlow with 20K iterations and a batch size of 32. For joint generalization, we follow previous RVC settings~\cite{2023uCFNet}, augmenting the Middlebury and ETH3D training sets to match KITTI 2015/2012 in size, preventing small datasets from being overwhelmed by large datasets. 
We then fine-tune the model on mixed datasets with 20K iterations.
All experiments are implemented in Pytorch using 8 RTX 5000 Ada GPUs. We use the AdamW~\cite{kingma2015adam} optimizer and a one-cycle learning rate schedule~\cite{smith2019super} with a maximum learning rate of $2\times10^{-4}$. For data augmentation, we employ asymmetric chromatic augmentations~\cite{li2022practical} and asymmetric occlusion~\cite{yang2019hierarchical,wang2025dualnet} to the right image. The image pairs are randomly cropped to \textbf{$320 \times 832$} for training.
We perform 16 iterations during training and report the results of 24 inference iterations.

\begin{table}[t]
\scriptsize
\caption{Domain generalization evaluation (peak results) on four target training sets. Note that, the inference time is evaluated on a KITTI stereo pair using a single Nvidia 5000 Ada GPU.}
\vspace{-0.2cm}
\setlength{\tabcolsep}{1.2mm}{
    \begin{tabular}{c|c|c|cccc}
    \toprule
    \multirow{2}{*}[-1pt]{\textbf{VFM}} & {Model} & Time & \multirow{1}{*}[-1pt]{\textbf{KIT 2012}}  &  \multirow{1}{*}[-1pt]{\textbf{KIT 2015}}  &  \multirow{1}{*}[-1pt]{\textbf{Middle}} &  \multirow{1}{*}[-1pt]{\textbf{ETH3D}}\\
    & Capacity & (s) & \textbf{Bad 3.0} & \textbf{Bad 3.0} & \textbf{Bad 2.0}  & \textbf{Bad 1.0}\\
    \midrule
    DAMV2 & ViT-Small~\cite{yang2024depth_v2} & 0.16 & 4.57 & 5.22 & 7.78 & 2.43 \\
    DAMV2 & ViT-Base~\cite{yang2024depth_v2} & 0.20 & \textcolor{blue}{\textbf{4.22}} & \textcolor{blue}{\textbf{4.86}} & \textcolor{blue}{\textbf{7.05}} & \textcolor{blue}{\textbf{2.10}} \\
    DAMV2 & ViT-Large~\cite{yang2024depth_v2} & 0.32 & \textbf{3.94} & \textbf{4.59} & \textbf{6.71} & \textbf{1.87}\\
    \bottomrule
    \end{tabular}}
\label{sec4:capacity}
\vspace{-0.2cm}
\end{table}

\begin{figure}[t]
    \centering
    \includegraphics[width=1\linewidth]{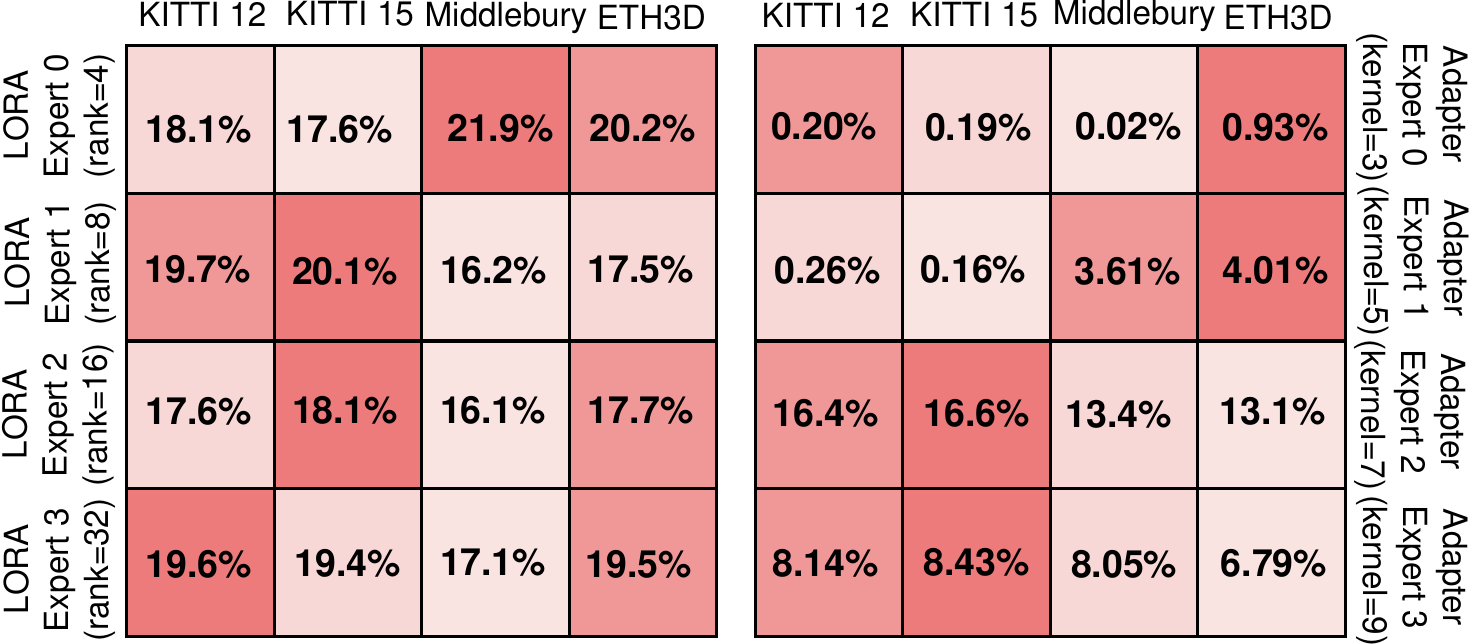}
    \vspace{-0.56cm}
    \caption{Dynamic expert selection on four real-world training datasets.
    Lighter colors represent lower activation rates.} 
    \label{sec4:matrix}
    \vspace{-0.3cm}
\end{figure}

\begin{table*}[h]
\centering
\scriptsize
\caption{Robustness comparison among ETH3D, Middlebury, and KITTI2015 testsets with existing SOTA methods in RVC.
All methods are tested on three datasets with a single fixed model. We evaluate Middlebury and ETH3D using the ``All" metrics.
The overall rank is obtained by Schulze Proportional
Ranking~\cite{schulze2011new} to combine multiple rankings into one. Our approach achieves the best overall performance.}
\vspace{-0.2cm}
\setlength{\tabcolsep}{0.8mm}{
    \begin{tabular}{l|cccc|cccc|cccc|c}
    \toprule
    \multirow{2}{*}[-1pt]{Methods} & 
    \multicolumn{4}{c|}{\textbf{Middlebury}} & 
    \multicolumn{4}{c|}{\textbf{KITIT 2015}} & \multicolumn{4}{c}{\textbf{ETH3D}} & Overall\\
    & \textbf{bad 2.0 (\%)} & \textbf{bad 4.0 (\%)}  & AvgErr & Rank & \textbf{D1-bg (\%)}  & \textbf{D1-fg (\%)}  & \textbf{D1-All (\%)} & Rank & \textbf{bad 1.0 (\%)} & \textbf{bad 2.0 (\%)} & AvgErr & Rank & Rank\\
    \midrule
    AANet\_RVC~\cite{xu2020aanet} & 31.8 & 25.8 &12.8 &11 &2.23 &4.89& 2.67& 9 &5.41& 1.95& 0.33 &9 &12 \\
    GANet\_RVC~\cite{zhang2019ga} & 24.9 & 16.3 & 15.8 & 10 &1.88& 4.58 &2.33 & 8 &6.97 &1.25& 0.45 & 11 & 11 \\
    HSMNet\_RVC~\cite{yang2019hierarchical} & 16.5 & 9.68 & 3.44 & 5 & 2.74 & 8.73 &3.74 & 10 & 4.40 & 1.51 & 0.28 & 8 & 10\\
    MaskLacGwcNet\_RVC~\cite{rao2023masked} & 15.8 & 10.3 & 13.5 &8 & 1.65 &3.68 &1.99& 6 &6.42 &1.88 & 0.38 &10 &9\\
    NLCANetV2\_RVC~\cite{rao2020nlca} &16.4 & 10.3 & 5.60& 8 &1.51 &3.97 &1.92 &4 &4.11 &1.20 &0.29 &7& 8\\
    CroCo\_RVC~\cite{weinzaepfel2023croco} &19.7 &12.2 &5.14 & 9 &2.04 &3.75 &2.33 &8 &1.54 &0.50 &0.21 & 3 & 7\\
    CFNet\_RVC~\cite{shen2021cfnet} &16.1 &11.3 &5.07 &7 &1.65 &3.53 &1.96 &5 &3.70& 0.97 &0.26& 6 &6\\
    UCFNet\_RVC~\cite{jiang2022improved} &16.7 & 10.9 &5.96 & 4 &1.57 &3.33 & 1.86 & 3 &3.37&0.78 &0.25 &5 &5\\
iRaftStereo\_RVC~\cite{jiang2022improved} &13.3 & 8.02 &2.90 & 4 & 1.88 &\textcolor{blue}{\textbf{3.03}}   &2.07& 7 &1.88 &0.55 &0.17 &4 &4\\
CREStereo++\_RVC~\cite{jing2023uncertainty} & \textcolor{blue}{\textbf{9.46}} &6.25 &\textcolor{blue}{\textbf{2.20}} & 3 &\textcolor{blue}{\textbf{1.55}}  &3.53 &1.88 & 4 &1.70 &0.37 &\textcolor{blue}{\textbf{0.16}} &2 &3 \\
LoS\_RVC~\cite{li2024local}  & \textbf{9.30} &\textcolor{blue}{\textbf{6.03}} &2.36& 2 &{1.58} &\textcolor{blue}{\textbf{3.03}}  &\textcolor{blue}{\textbf{1.83}} &2 &\textcolor{blue}{\textbf{1.47}} & \textbf{0.25} & \textbf{0.14} & 1 & 2 \\
\midrule
\textbf{SMoEStereo\_RVC (Ours)} & 9.74 & \textbf{5.41} & \textbf{1.94} & 1 & \textbf{1.50} & \textbf{2.88} & \textbf{1.73} & 1 & \textbf{1.13} & \textcolor{blue}{\textbf{0.26}} & \textbf{0.14} & \textbf{1} & 1\\
    \bottomrule
    \end{tabular}}
    \vspace{-0.3cm}
\label{sec4:comparison}
\end{table*}

\begin{figure*}[h]
    \centering
    \includegraphics[width=1\linewidth]{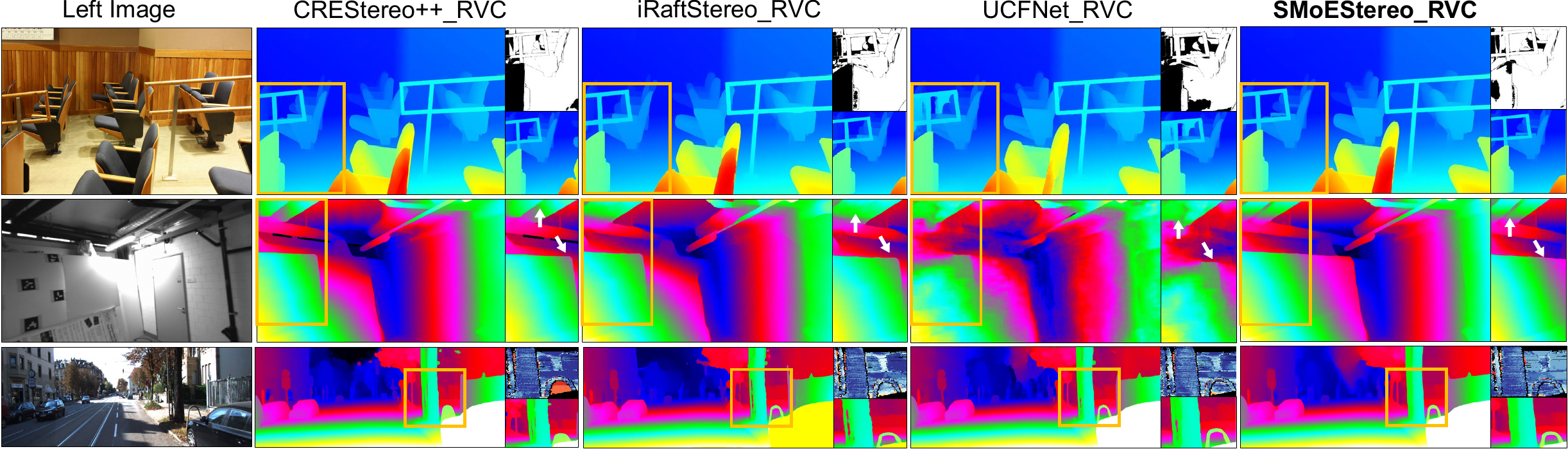}
    \vspace{-0.6cm}
    \caption{Qualitative results on the test set of Middlebury, ETH3D, and KITTI2015. Since ETH3D benchmark do not provide
the error map, we only show the zoomed highlight parts.
In the error map, lighter colors represent smaller errors, and vice versa.
 Best zoomed in.}
\label{sec4:visual_comparison}
\vspace{-0.4cm}
\end{figure*}

\subsection{Robustness Evaluation}

In stereo matching, cross-domain generalization~\cite{zhang2020domain,zhanglearning2024} and joint generalization~\cite{2023uCFNet,shen2021cfnet} are key for evaluating the robustness of stereo matching methods.
Therefore, we conduct these two types of robustness experiments.

\vspace{-0.1cm}
\subsubsection{Cross-domain Generalization}
Our SMoEStereo first demonstrates strong robustness in zero-shot settings, as evidenced in Table~\ref{sec4:cross_comparison} and Table~\ref{sec4:weather_comparison}.

\noindent \textbf{Comparison with Domain Generalized Approaches.}
Our SMoEStereo significantly outperforms most stereo matching methods tailored for domain generalization. Notably, HVT-RAFT~\cite{chang2023domain}, Former-PSMNet, Former-CFNet, and Former-RAFT~\cite{zhanglearning2024} use specialized modules and losses for domain generalization, while our method does not require. 
We also compare the robustness of SMoEStereo under challenging outdoor conditions such as cloud, fog, and rain, as illustrated in Table~\ref{sec4:weather_comparison}.
Additionally, while Former-RAFT~\cite{zhanglearning2024} uses the ViT-Large of DAM, we employ the smaller ViT-Base version. As shown in Table~\ref{sec4:vfm_comparison}, our method has fewer parameters and faster inference time . 

\noindent \textbf{More VFMs.} We extend our experiments by integrating SMoE with other VFMs, such as DAM~\cite{depthanything}, SAM~\cite{kirillov2023segment}, DINOV2~\cite{oquab2024dinov2} to highlight the versatility of our SMoE. 
As shown in Table~\ref{sec4:cross_comparison}, our findings reveal that SMoE exhibits remarkable performance with diverse VFM backbones.

\noindent \textbf{Comparing SMoE with PEFT Methods.}
We conduct a comprehensive performance comparison of SMoE against existing PEFT methods for domain generation, as detailed in Table~\ref{sec4:peft}.
In addition to `Frozen' and `Full-finetuning', we developed four network variants by replacing SMoE modules with other PEFT methods: VPT~\cite{jia2022visual}, Adapter~\cite{chen2022adaptformer}, LoRA~\cite{hu2021lora}, and Adapter-Tuning~\cite{zhanglearning2024}. 
Leveraging the robust feature extraction of VFMs, these PEFT methods have shown notable advancements in generalization ability. Using identical VFM backbones, SMoE outperforms previous domain generalization and other PEFT methods. Additionally, SMoE enjoys fewer parameters during the inference phase than other PEFT methods.

\noindent \textbf{VFM Capacity.}
Using DAMV2~\cite{yang2024depth_v2} as an example, we develop variants with different capacities to assess the impact on zero-shot performance. Table~\ref{sec4:capacity} shows that larger capacities enhance zero-shot performance, as larger models provide stronger representation abilities and robust priors essential for generalization. 

\noindent \textbf{Expert Selection Distribution.}
\textcolor{black}{Different stereo datasets exhibit significant domain gaps. From Fig.~\ref{sec4:matrix}, we observe distinct LoRA and Adapter expert selection distributions among these four datasets. 
Our SMoE framework dynamically activates the optimal combination of LoRA and Adapter experts for each dataset, which empirically validates SMoE’s flexible adaptability - a critical advantage for in-the-wild deployment, where robust cross-domain generalization hinges on dynamic expert selection.}

\begin{table*}[t]
\caption{Ablation of the proposed method on Middlebury, KITTI2015, and ETH3D training datasets. 
DAMV2~\cite{yang2024depth_v2} (ViT-Base) is used as the frozen VFM. 
Our MoE LORA and MoE Adapter selectively activate the optimal expert rather than all experts.
Params. denote the extra activated parameters within the DAMV2 backbone. \textcolor{Apricot}{Apricot color} represents the baseline results, while \textcolor{SkyBlue}{blue color} represents the final model results.
The network component is evaluated individually in the table. Inference time is measured on KITTI by A5000 Ada GPU.}
\centering
\scriptsize
\vspace{-0.3cm}
\setlength{\tabcolsep}{2.5mm}{
    \begin{tabular}{c|cccc|cc|cc|cc|ccc}
    \toprule
      \multirow{2}{*}{ID} & MoE & MoE & Decision  & Random  & \multicolumn{2}{c|}{\textbf{KITTI 2015}}  &  \multicolumn{2}{c|}{\textbf{Middlebury}} & \multicolumn{2}{c|}{\textbf{ETH3D}}  & {Number of} & {Params.} & Runtime \\
    &  LORA & Adapter & Network & Policy & \textbf{EPE} & \textbf{D1\_All}  & \textbf{EPE} & \textbf{Bad 2.0}  & \textbf{EPE}  & \textbf{Bad 1.0}  & MoE & (M) & (s)\\
    \midrule
       1 & -& - & -  & - & \cellcolor{Apricot} 0.72  &  \cellcolor{Apricot} 2.01 &  \cellcolor{Apricot} 0.97 & \cellcolor{Apricot} 7.67 & \cellcolor{Apricot} 0.25 & \cellcolor{Apricot} 1.79  & \cellcolor{Apricot} 0 & \cellcolor{Apricot} 0 & \cellcolor{Apricot} 0.17 \\
   2 & $\checkmark$ & - & -  & - & 0.63 & 1.65 & 0.81 & 4.80 & 0.17 & 0.92 & 12 & 2.29 & 0.19 \\
   3 & - &$\checkmark$ & -  & - & 0.64 & 1.71 & 0.77 & 4.68 & 0.18 & 1.01 & 12 & 4.49 & 0.19 \\
   4 & $\checkmark$ & $\checkmark$ & - & - &  {0.61} & {1.57} & {0.72} & {4.24} & \textbf{0.14} & \textbf{0.56} & 24 & 6.77 &  0.23 \\
   5 & $\checkmark$ & $\checkmark$ & - & $\checkmark$ & 0.64 & 1.69 & 0.75 & 4.66 & 0.17 & 0.84 & $\sim 14$ & 2.86 & 0.20 \\
   6 & $\checkmark$ & $\checkmark$ & $\checkmark$  & - & \cellcolor{SkyBlue} \textbf{0.60} & \cellcolor{SkyBlue} \textbf{1.51} & \cellcolor{SkyBlue} \textbf{0.71} & \cellcolor{SkyBlue} \textbf{4.12} & \cellcolor{SkyBlue}  \cellcolor{SkyBlue}  {0.15} & \cellcolor{SkyBlue}  0.63 & \cellcolor{SkyBlue} 
 $\sim$\textbf{14} & \cellcolor{SkyBlue} 
 \textbf{2.86} & \cellcolor{SkyBlue}  \textbf{0.20} \\
    \bottomrule
    \end{tabular}}
    \vspace{-0.5cm}
\label{sec4:ablation}
\end{table*}

\vspace{-0.1cm}
\subsubsection{Joint Generalization Evaluation}

Table~\ref{sec4:comparison} presents the results of our method and existing SOTA methods in stereo matching for the Robust Vision Challenge (RVC). 
In RVC, all methods should be jointly trained on real-world datasets (KITTI 2012 \& 2015, Middlebury, and ETH3D) and evaluated on three public benchmarks using a single fixed model without adaptation. 
As shown, SMoEStereo\_RVC achieves 1$^{st}$ on these public benchmarks among all methods, clearly demonstrating the advantage of our approach. 
Note that, LoS\_RVC~\cite{likh2024los} use additional large-scale datasets such as Instereo2K~\cite{bao2020instereo2k} and CRE~\cite{li2022practical} to achieve impressive results.  
The larger data capacity improves the overall performance (See supplementary).
The visual comparison results are shown in Fig.~\ref{sec4:visual_comparison}.

\subsection{Ablation Studies}
We conduct ablation studies to demonstrate the efficacy of each network component in RVC settings. Cross-domain generalization ablations are detailed in the Supplementary.

\begin{table}[]
    \centering
    \scriptsize
    \caption{Impacts of the matrix rank of MoE LoRA layers.}
    \vspace{-0.3cm}
    \setlength{\tabcolsep}{2.5mm}{
    \begin{tabular}{c|cc|cc|cc}
    \toprule
     \multirow{2}{*}[-1pt]{Setting}  & \multicolumn{2}{c|}{\textbf{KITTI 2015}}  &  \multicolumn{2}{c|}{\textbf{Middlebury}} & \multicolumn{2}{c}{\textbf{ETH3D}} \\
    & \textbf{EPE} & \textbf{D1\_All}  & \textbf{EPE} & \textbf{Bad 2.0}  & \textbf{EPE}  & \textbf{Bad 1.0} \\
    \midrule
    rank = 4 & 0.65 & 1.77 & 0.85 & 4.75 &  \textcolor{blue}{\textbf{0.16}} & \textcolor{blue}{\textbf{0.73}} \\
    rank = 8 & 0.64 & 1.73 & 0.78 & 4.64 & 0.17 & 0.80 \\
    rank = 16 & 0.63 & 1.61 & \textcolor{blue}{\textbf{0.74}} & \textcolor{blue}{\textbf{4.29}} & 0.18 & 0.84 \\
    rank = 32 & \textcolor{blue}{\textbf{0.62}} & \textcolor{blue}{\textbf{1.57}} & 0.76 & 4.42 & \textcolor{blue}{\textbf{0.16}} & 0.77 \\
    \midrule
    SMoE  & \textbf{0.60} & \textbf{1.51} & \textbf{0.71} & \textbf{4.12} & \textbf{0.15} & \textbf{0.63}  \\
    \bottomrule
    \end{tabular}}
    \label{sec4:rank}
    \vspace{-0.25cm}
\end{table}

\begin{table}[]
    \centering
    \scriptsize
    \caption{SMoE \textit{vs.} Multi-E (All experts are involved in MoE).}
    \vspace{-0.3cm}
    \setlength{\tabcolsep}{0.4mm}{
    \begin{tabular}{c|cc|c|cc|cc|cc}
    \toprule
    \multirow{2}{*}[-1pt]{Setting} & {Training} & {Inference} & {Number} & \multicolumn{2}{c|}{\textbf{KITTI 2015}}  &  \multicolumn{2}{c|}{\textbf{Middlebury}} & \multicolumn{2}{c}{\textbf{ETH3D}} \\
    & Time (s) & Time (s) & of MoE & \textbf{EPE} & \textbf{D1\_All}  & \textbf{EPE} & \textbf{Bad 2.0}  & \textbf{EPE}  & \textbf{Bad 1.0} \\
    \midrule
    Multi-E & 2.10 iter/s & 4.68 iter/s & 24 & \textbf{0.59} & \textbf{1.46} & 0.76 & 4.37 & 0.17 & 0.78  \\
    \midrule
    SMoE  & \textbf{2.63 iter/s} & \textbf{5.26 iter/s} & 14 & 0.60 & 1.51 & \textbf{0.71} & \textbf{4.12} & \textbf{0.15} & \textbf{0.63} \\
    \bottomrule
    \end{tabular}}
    \label{sec4:moe}
    \vspace{-0.6cm}
\end{table}

\noindent \textbf{Ablation of Main Components.}
To evaluate the effectiveness of the designed PEFT modules, we report cross-dataset accuracy results in Table~\ref{sec4:ablation}. Utilizing MoE LoRA and MoE adapters enhances disparity estimation performance compared to the vanilla VFM baseline (ID = 1) by capturing long-range interactions and local geometry cues. 
This improvement is attributed to our MoE design, which effectively learns robust features in the target domain while preserving knowledge from dense prediction tasks.
However, with more MoE modules, computational costs inevitably increase (ID = 4). The proposed decision network mitigates this by removing redundancy while maintaining high performance (ID = 6). Additionally, replacing learned usage policies with randomly generated ones of similar computational cost (ID = 5) results in a notable drop in accuracy, confirming the effectiveness of the learned usage policy.

\begin{figure}
    \centering
\includegraphics[width=1\linewidth]{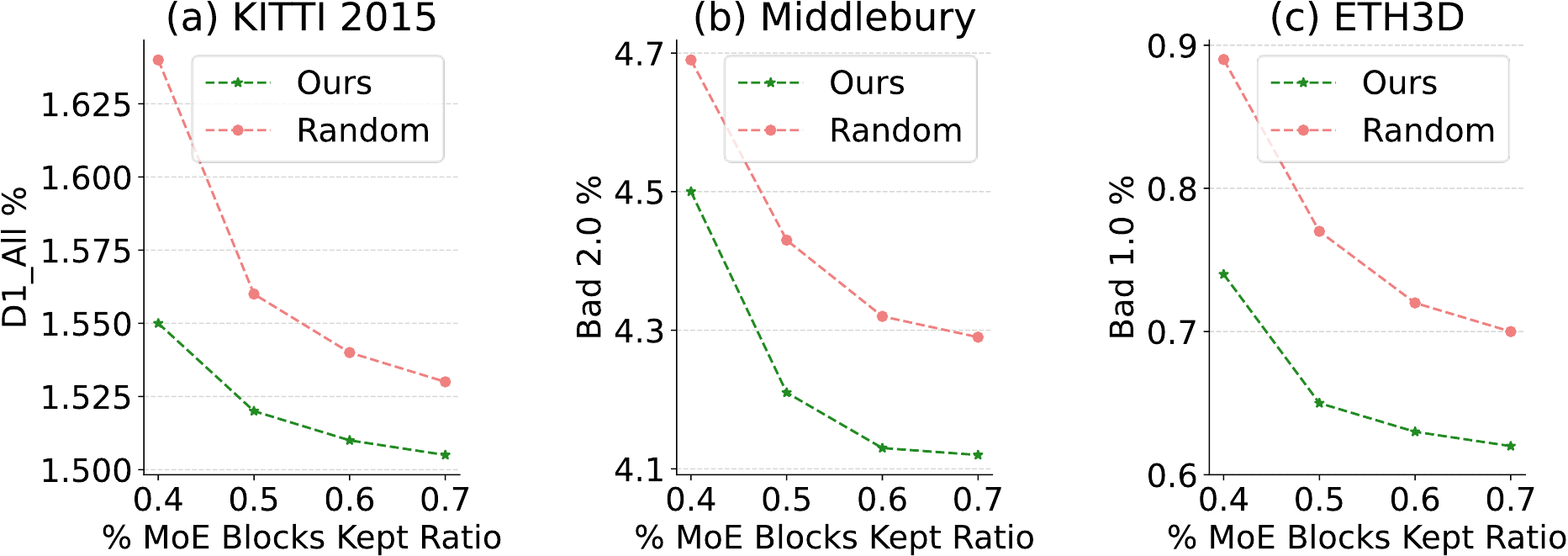}
    \vspace{-0.7cm}
    \caption{Quantitative results on real-world datasets between the proposed layer selection strategy and their Random counterparts.}
    \vspace{-0.65cm}
    \label{sec4:chart}
\end{figure}

\noindent \textbf{Study on Rank $r$.}
In Table~\ref{sec4:rank}, we compare the cross-dataset performance of SMoEStereo with LoRA experts of varying ranks. Different datasets exhibit varying optimal LoRA ranks. Our heterogeneous MoE dynamically selects the optimal LoRA rank for each input query, consistently outperforming the homogeneous MoE LoRA experts on all three datasets. Similarly, the impact of different adapters and LoRA with more ranks is investigated in the supplementary.
This fully demonstrates the effectiveness of our scene-conditioned selection mechanism in stereo matching.

\noindent \textbf{SMoE \textit{vs.} Multi Experts.}
To demonstrate the effectiveness of SMoE's dynamic selection of optimal LoRA and Adapter experts, we compare it with Multi Experts (Multi-E), which aggregates outputs from all designed experts. As shown in Table~\ref{sec4:moe}, the SMoE mechanism selectively activates sparse experts, achieving a $1.25\times$ speedup in training and a $1.12\times$ speedup in inference. Despite these gains, SMoE outperforms Multi-E in cross-dataset performance, highlighting its effectiveness in selecting the optimal expert for robust stereo matching. This suggests that naively aggregating multiple experts may not be optimal, as it can suppress informative features while introducing noise.

\noindent \textbf{Different Computational Budgets.}
SMoEStereo is designed to accommodate the needs of different computational budgets flexibly by varying the hyperparameter $\gamma$ (MoE kept ratio). As demonstrated
in Fig.~\ref{sec4:chart}, our method can cover a wide range of trade-offs between efficiency and accuracy and outperforms Random baselines significantly.

\section{Conclusion}
\label{sec5}
In this paper, we introduce SMoEStereo, a general framework designed to effortlessly leverage VFMs for robust stereo matching in the wild. 
By dynamically selecting suitable MoE and experts based on inputs, SMoEStereo exhibits strong robustness with few learnable parameters, significantly outperforming previous robust methods.
Experimental results show that our approach performs well on various datasets and has generic applicability. 

{
    \small
    \bibliographystyle{ieeenat_fullname}
    \bibliography{main}
}


\end{document}


\clearpage
\setcounter{page}{1}
\maketitlesupplementary

\section{Overview}
In this supplementary material, we provide additional details of our method and experiments, including:

\begin{itemize}
\item  Analysis on runtime efficiency~\cite{zhanglearning2024}.
\end{itemize}

\begin{itemize}
\item The detailed experimental settings.
\end{itemize}

\begin{itemize}
\item More ablation studies.
\end{itemize}

\begin{itemize}
\item The data capacity of SMoEStereo when more synthetic samples are used for training.
\end{itemize}

\begin{itemize}
\item More zero-shot visualization results.
\end{itemize}

\begin{table}[t]
\centering
\caption{Runtime efficiency comparison on KITTI 2015.}
\scriptsize
\vspace{-0.5em}
\label{supp:efficiency}
    \setlength{\tabcolsep}{1mm}{
    \begin{tabular}{*{10}{c}}
        \toprule
      Method  & Feature Extractor & Iterations & Time (\textit{s}) & Res. & GPU \\
        \midrule
        RAFTStereo~\cite{Lipson2021RAFTStereoMR}  & ResNet-like & 32 & 0.17 & 1242 x 375  & 5000Ada \\
       DLNR~\cite{zhao2023high}  & Transformer & 32& 0.27  & 1242 x 375  & 5000Ada \\
        Selective-IGEV~\cite{wang2024selective} & ResNet-like & 32 & 0.21 & 1242 x 375 & 5000Ada \\
        MochaStereo~\cite{chen2024mocha} & ResNet-like & 32  & 0.28 & 1242 x 375 & 5000Ada \\
        Former-RAFT~\cite{zhanglearning2024} & ViT-Large & 32 & 0.47 & 1242 x 375 & 5000Ada \\
        \midrule
        \textbf{Ours}  & ViT-Base & 24 & 0.20  & 1242 x 375 & 5000Ada \\
        \bottomrule
    \end{tabular}}
\end{table}

\section{Analysis on Runtime Efficiency} 
Our SMoEStereo achieves better inference efficiency than most RAFT-based methods, as shown in Tab.~\ref{supp:efficiency}. Although VFMs slightly increase feature extraction time, their robust features significantly reduce the need for iterative disparity refinements. Specifically, our GRU performs only 24 iterations \textit{vs.} 32 iterations in ~\cite{Lipson2021RAFTStereoMR,zhao2023high,wang2024selective,zhao2023high,zhanglearning2024}, thus improving overall efficiency.
Notably, SMoE offers a flexible selection mechanism to control the number of MoE modules, reducing inference time for various real-world applications. We benchmark inference runtime on KITTI 2015 (1242×375) using an RTX 5000 Ada GPU.

\section{More Details about Experiments}
\subsection{More Details about VFMs.}
\noindent\textbf{SAM.} Aligning with the methodology described in the foundational paper~\cite{kirillov2023segment}, we employ the ViT-Large architecture as our image encoder, making use of pre-trained weights that were trained on SA-1B~\cite{kirillov2023segment} for a promptable segmentation task. The patch size of this model is set to 16×16, and each layer is designed to output features with a dimensionality of 768, summing up to a total of 12 layers.
The positional embeddings of the model are upscaled to a length of 1024 via bicubic interpolation. From this model, we extract features from the 2nd, 5th, 7th, and 11th layers layers and feed them into the decoder.

\noindent\textbf{DAM\& DAMV2.} DAM and DAMV2 designed a data engine to automatically generate depth annotations for unlabeled images, enabling data scaling up to an arbitrary scale. It collects 62M diverse and informative images from eight public large-scale datasets, e.g., SA-1B~\cite{kirillov2023segment}, Open Images~\cite{kuznetsova2020open}, and BDD100K~\cite{yu2020bdd100k} for training an initial MDE mode in a self-training manner~\cite{lee2013pseudo}.
Similar to SAM, we also employ the ViT-Base capacity as our image encoder.
The patch size of this model is set to 16×16, and each layer is designed to output features with a dimensionality of 128, summing up to 12 layers.
The positional embeddings of the model are upscaled to a length of 1024 via bicubic interpolation. From this model, we extract features from the 2nd, 5th, 8th, and 11th layers and feed them into the subsequent cost aggregators.

\noindent\textbf{DINOv2.} Our choice of backbone for this study is DINOv2-Base, which has been distilled from DINOv2-Large. As noted in the original documentation, DINOv2-Base occasionally surpasses the performance of DINOv2-Large~\cite{oquab2024dinov2}. 
we apply equivalent processing to both the positional embeddings and patch embed layer of DINOv2-Base.
The features extracted from the 2nd, 5th, 8th, and 11th layers are subsequently fed into the decode head. 
DINOv2 is originally pretrained in a self-supervised fashion on the LVD-142M~\cite{oquab2024dinov2} dataset, following the procedures outlined in its respective paper.

\subsection{More Details about PEFT Methods.}
\noindent\textbf{1. VPT, Adapter-Tuning, AdaptFormer, and LoRA}. 
Based on extensive experimentation, we have optimized the implementation of PEFT methods for DAMV2 and SAM, utilizing PEFT configurations that enhance robust performance. 
These methods include: 
1) VPT: We use the VPT-Deep configuration and incorporate 256 learnable tokens within each ViT layer (12 layers for ViT-Base).

2) LoRA: Applied to the query and value components for self-attention, default configured with a rank of 128.

3) AdaptFormer: Similar to LoRA, equipped with MLP layers and employs a bottleneck design with a default ank of 32, initialized using LoRA, and notably omits layer normalization.

4) AdapterTuning: We adopt a DPT decoder 
strategy~\cite{Ranftl2021} with multi-scale fusion, utilizing input channels of 128 dimensions.

\noindent\textbf{2. Frozen, Full Finetuning, and LoRA}. 
We define the differences between these fine-tuning methods.

\textbf{Frozen}: The VFM backbone is frozen and SMoE modules are removed. Only the shallow CNN (Fig.2a) and subsequent GRU modules are trainable.

\textbf{Full Finetuning}: All components are trainable.

\textbf{LoRA}: The single LoRA layer (fixed rank=128) is used within each ViT block.
 Besides, Tab.~\ref{supp:lora} explores varying ranks to demonstrate SMoE’s dynamic design effectiveness.

\begin{table}[t]
    \centering
    \scriptsize
    \caption{Zero-shot Non-Lambertian Generalization. Comparison with state-of-the-art models. Networks trained on SceneFlow. We use the officially provided weights.}
    \scriptsize
    \vspace{-0.3cm}
    \setlength{\tabcolsep}{2mm}{
    \begin{tabular}{c|ccccc}
    \toprule
    \multirow{2}{*}[-1pt]{Model} & \textbf{\textgreater 2 px}   & \textbf{\textgreater 4 px} & \textbf{\textgreater 6 px}  &  \textbf{\textgreater 8 px} & \textbf{Avg.} \\
    & \textbf{(\%)} & \textbf{(\%)} &  \textbf{(\%)} & \textbf{(\%)} & \textbf{(px)} \\
    \midrule
    PSMNet~\cite{chang2018pyramid} & 34.5 & 24.8 & 20.5 & 17.8 & 7.30 \\
    RAFTStereo~\cite{Lipson2021RAFTStereoMR} & 17.8 & 13.1 & 10.8 & 9.24 & 3.60 \\
    Selective-RAFT~\cite{wang2024selective} & 20.0 & 15.1  & 12.5 & 10.9  & 4.12 \\
   Selective-IGEV~\cite{wang2024selective} & 18.5 & 14.2 & 12.1 & 10.8 & 4.38 \\
   DLNR~\cite{zhao2023high} & 18.6 & 14.6 & 12.6 & 11.2 & 3.97 \\
   \midrule
   \textbf{SMoEStereo (Ours)} & \textbf{11.3} & \textbf{7.16} & \textbf{6.47} & \textbf{5.13} & \textbf{2.09} \\
    \bottomrule
    \end{tabular}}
    \label{supp:tab_non}
\end{table}

\section{More Ablation Studies}
In this section, we systematically evaluate the components of our framework through four key analyses. First, we conduct cross-domain generalization ablations on SceneFlow, dissecting the contribution of each SMoE component to domain-shift robustness. Next, we investigate architectural compatibility, validating SMoE’s integration with diverse robust training frameworks (i.e., DKT framework~\cite{zhang2024robust}) and classical stereo architectures (e.g., IGEVStereo~\cite{xu2023iterative}, PSMNet~\cite{chang2018pyramid}). Additionally, We further analyze the computational efficiency of MoE LoRA and MoE Adapters by profiling their dynamic token allocation patterns across network layers, quantifying parameter savings under varying scene complexities. Concurrently, we evaluate the effectiveness of MoE balance losses in ensuring equitable expert utilization, demonstrating their role in stabilizing training while maintaining task performance.
Finally, we assess data scalability by measuring performance gains from incremental synthetic data integration.

\begin{table*}[t]
\caption{Cross-domain performance ablation study trained on SceneFlow. DAMV2 (ViT-Base) used.}
\label{supp:abl_sc}
\vspace{-0.3cm}
\centering
\scriptsize
    \begin{tabular}{*{14}{c}}
    \toprule
      \multirow{2}{*}{ID} & \textbf{MoE} & \textbf{MoE}  & \textbf{Decision} & \textbf{Random} & \multicolumn{1}{c}{{\textbf{KITTI 2012}}} & \multicolumn{1}{c}{{\textbf{KITTI 2015}}}  &  \multicolumn{1}{c}{{\textbf{Middlebury}}} & \multicolumn{1}{c}{{\textbf{ETH3D}}}  & {\textbf{Number of}} & {\textbf{Used Params.}} \\
   &  \textbf{LORA} & \textbf{Adapter} & \textbf{Network} & \textbf{Decision} & \textbf{{D1\_All}} & \textbf{{D1\_All}} & \textbf{{Bad 2.0}}  & \textbf{{Bad 1.0}}  & \textbf{MoE} & \textbf{(M)} \\
    \midrule
    1 & - & - & - & - & \cellcolor{Apricot} 11.7  & \cellcolor{Apricot} 15.4  & \cellcolor{Apricot} 24.6 & \cellcolor{Apricot} 16.2 & \cellcolor{Apricot} 0 & \cellcolor{Apricot} 0 \\
   2 & $\checkmark$ & - & - & - & 4.31  & 4.93 & 7.60 & 2.56 & 12 & 2.29 \\
   3 & - &$\checkmark$ & -  & - & 4.45 & 5.03 & 7.43 & 2.41 & 12 & 4.49 \\
   4 & $\checkmark$ & $\checkmark$ & - & - & \textbf{4.19} & \textbf{4.79} & \underline{7.14} & \textbf{1.99} & 24 & 6.77 \\
   5 & $\checkmark$ & $\checkmark$ & - & $\checkmark$ & {4.51} & 5.19 & 8.32 & 3.04 & ~14 & 2.86 \\
   6 & $\checkmark$ & $\checkmark$ & $\checkmark$ & - & \cellcolor{Lavender} \underline{4.22} & \cellcolor{Lavender} \underline{4.86} & \cellcolor{Lavender} \textbf{7.05}  & \cellcolor{Lavender} \underline{2.10} & \cellcolor{Lavender} ~14 & \cellcolor{Lavender} 2.86 \\
    \bottomrule
    \end{tabular}
\end{table*}

\begin{table}[t]
\caption{Ablations of the dynamic selection of MoE LoRA layers.}
\vspace{-0.2cm}
\label{sec4:tab_moe_lora}
\centering
\scriptsize
    \setlength{\tabcolsep}{1mm}{
    \begin{tabular}{*{9}{c}}
    \toprule
    \multirow{2}{*}{\textbf{Setting}} & \multirow{2}{*}{\textbf{VFM}} & \textbf{KIT 2012}  &  \textbf{KIT 2015}  &  \multicolumn{1}{c}{{\textbf{Middle}}}  & \multicolumn{1}{c}{{\textbf{ETH3D}}}  \\
    & & \textbf{Bad 3.0} &  \textbf{Bad 3.0} &  \textbf{Bad 2.0} &  \textbf{Bad 1.0} \\
    \midrule
    MoE LoRA (Rank = 4) & DAMV2 & 4.55 & 5.01 & 7.93 & 3.08 \\ 
    MoE LoRA (Rank = 8) & DAMV2 & 4.60 & 5.11 & 8.01 & 2.93 \\ 
    MoE LoRA (Rank = 16) & DAMV2 & 4.74 & 5.17 & 7.82 & 3.39 \\ 
    MoE LoRA (Rank = 32) & DAMV2 & 4.37 & 4.97 & 8.10 & 2.87 \\ 
    \midrule
    \textbf{SMoE w.o/ MoE Adapter} & DAMV2 &  \textcolor{blue}{\textbf{{4.31}}} &\textcolor{blue}{\textbf{{4.93}}} & \textcolor{blue}{\textbf{{7.60}}} & \textcolor{blue}{\textbf{{2.56}}}  \\
    \textbf{SMoE} & DAMV2 & \textbf{4.22} & \textbf{4.86} & \textbf{7.05} & \textbf{2.10}  \\
    \bottomrule
    \end{tabular}}
    \vspace{-0.2cm}
\end{table}

\begin{table}[t]
\centering
\scriptsize
\caption{Zero-shot performance comparison of the proposed SMoE against other finetuning methods. Params (M) Train/Test refers to the learnable and additional activated parameters within the VFM backbone for the training and inference phases, respectively.}
\vspace{-0.3cm}
\setlength{\tabcolsep}{0.05mm}{
    \begin{tabular}{c|c|c|c|c|c|c}
    \toprule
    \multirow{2}{*}[-1pt]{Backbone} & {Fine-tune} & { Params (M)} & \multirow{1}{*}[-1pt]{\textbf{KIT 2012}} & \multirow{1}{*}[-1pt]{\textbf{KIT 2015}}  &  \multirow{1}{*}[-1pt]{\textbf{Middle}} &  \multirow{1}{*}[-1pt]{\textbf{ETH3D}}\\
    & Method & Train/Test  & \textbf{Bad 3.0} & \textbf{Bad 3.0} & \textbf{Bad 2.0}  & \textbf{Bad 1.0}\\
    \midrule
    \multirow{6}{*}[-1pt]{DAMV2~\cite{yang2024depth_v2}} & LoRA~\cite{hu2021lora} (rank=4) & 0.15/0.15 & 4.62 & 5.31 & 7.92 & 3.12 \\
     & LoRA~\cite{hu2021lora} (rank=8) & 0.30/0.30 & \textcolor{blue}{\textbf{4.44}} & 5.07 & 8.22 & 3.08 \\
      & LoRA~\cite{hu2021lora} (rank=16) & 0.59/0.59 & 4.81 & 5.43 & \textcolor{blue}{\textbf{7.64}} & 2.87 \\
    & LoRA~\cite{hu2021lora} (rank=32) & 1.18/1.18 & 4.56 & {5.21} & 8.43 & \textcolor{blue}{\textbf{2.79}}\\ 
      (ViT-base) &  LoRA~\cite{hu2021lora} (rank=64) & 2.36/2.36 & 4.51 & 5.19 & 7.92 & 2.94 \\
      & LoRA~\cite{hu2021lora} (rank=128) & 4.72/4.72 & 4.47 & \textcolor{blue}{\textbf{5.03}} & 7.67 & 2.83 \\
    & \textbf{SMoE (Ours)} & 6.81/2.86 & $\textbf{4.22}$& $\textbf{4.86}$ & $\textbf{7.05}$ & \textbf{2.10} \\
    \midrule
    \multirow{6}{*}[-1pt]{SAM~\cite{kirillov2023segment}} & LoRA~\cite{hu2021lora} (rank=4) & 0.15/0.15 & \textcolor{blue}{\textbf{4.41}} & 5.04 &7.85 & 2.99 \\
     & LoRA~\cite{hu2021lora} (rank=8)& 0.30/0.30 & 4.63 & 5.30 & \textcolor{blue}{\textbf{7.68}} & 3.02 \\
      & LoRA~\cite{hu2021lora} (rank=16) & 0.59/0.59  & 4.55  & 5.19 & 7.98 & 2.81  \\
   & LoRA~\cite{hu2021lora} (rank=32) & 1.18/1.18 & 4.48 & \textcolor{blue}{\textbf{4.98}} & 8.35 & 2.76 \\ 
    (ViT-base) & LoRA~\cite{hu2021lora} (rank=64) & 2.36/2.36 & 4.60  & 5.23 & 8.02 & \textcolor{blue}{\textbf{2.66}} \\
     & LoRA~\cite{hu2021lora} (rank=128) & 4.72/4.72 & 4.49 & 5.17 & 7.76 & 2.78 \\
    & \textbf{SMoE (Ours)} & 6.81/4.06 & \textbf{4.27} & \textbf{4.89} & \textbf{7.10} & \textbf{2.07} \\
    \midrule
    \end{tabular}}
\label{supp:lora}
\vspace{-0.4cm}
\end{table}

\noindent \textbf{Ablation of Main Components.}
Table~\ref{supp:abl_sc} presents the cross-domain generalization accuracy results, demonstrating that the integration of MoE LoRA and MoE Adapter layers significantly enhances disparity estimation performance compared to the vanilla VFM baseline (ID = 1). This improvement can be attributed to the inherent capability of our MoE architecture to adaptively learn domain-invariant features while preserving the transferable knowledge acquired from dense prediction tasks, thereby fostering robust generalization across diverse scenarios. However, the increased computational overhead associated with additional MoE modules (ID = 4) highlights a trade-off between performance and efficiency. To address this, our proposed decision network effectively reduces redundancy by selectively activating the most relevant MoE components, achieving a balance between computational efficiency and high performance (ID = 6). Notably, replacing the learned usage policy with a randomly generated one of comparable computational cost (ID = 5) results in a significant decline in accuracy, underscoring the critical role of the learned policy in optimizing expert selection and ensuring superior cross-domain adaptability.

\noindent \textbf{Dynamic Expert Selection Mechanism.}
We conduct a comprehensive performance evaluation of SMoE across varying ranks of Low-Rank Adaptation (LoRA) for domain generalization, as detailed in Table~\ref{sec4:tab_moe_lora}. The zero-shot performance exhibits notable variability across different target domains, highlighting the critical influence of LoRA rank selection on model adaptability and generalization capabilities. This variability underscores the necessity of dynamically optimizing rank configurations to tailor the model’s representational capacity to the unique characteristics of each domain, thereby maximizing cross-domain performance and ensuring robust generalization in diverse scenarios. Overall, this fully demonstrates the effectiveness of dynamic expert selection mechanisms with varying ranks.

\noindent \textbf{Comparing SMoE with Different Rank of LoRA.}
We conduct a comprehensive performance comparison of SMoE against Low-Rank Adaptation (LoRA) models with varying ranks (e.g., r=4, 8, 32, 64, 128) for cross-domain generation tasks, as detailed in Table~\ref{supp:lora}. Our analysis reveals stark variability in zero-shot generalization across distinct target domains (e.g., indoor and outdoor scenes)
For various target domains, the zero-shot performance demonstrates significant variability, influenced by the specific ranks of Low-Rank Adaptation (LoRA) models. This variability underscores the importance of selecting appropriate ranks to optimize performance across different domains.
Naively adopting the single uniform LoRA is inadequate for robust stereo matching.

\noindent \textbf{ DKT framework~\cite{zhang2024robust} on our SMoEStereo.} 
DKT~\cite{zhang2024robust} is a robust training framework that was trained on real-world target datasets and demonstrates strong generalization capabilities across diverse datasets. In this section, we provide a comprehensive experimental analysis of the DKT framework, highlighting its strengths and limitations.
Specifically, under identical DKT settings, our method achieves significant performance improvements over DKT-RAFT, as evidenced in Tab.~\ref{tab:4}. These results underscore the effectiveness of our proposed Sparse Mixture of Experts (SMoE) design, demonstrating its ability to enhance robustness and adaptability in challenging scenarios.

\begin{table*}[htb]
\centering
\caption{Comparison with DKT-RAFT trained on KITTI.}
\label{tab:4}
    \small
    \begin{tabular}{*{12}{c}}
        \toprule
       \multirow{2}{*}[-1pt]{Method}  &  \multicolumn{2}{c}{KIT 2012} &  \multicolumn{2}{c}{KIT 2015} & \multicolumn{5}{c}{DrivingStereo} & MIDDLE & ETH3D \\
       & Noc & All & Bg & All & Sunny & Cloudy & Rainy & Foggy & Avg. & (Bad 2.0)  & (Bad 1.0) \\
        \midrule
        DKT-RAFT~\cite{zhang2024robust}  &  \cellcolor{pink} 1.43 &  \cellcolor{pink} 1.85 &  \cellcolor{pink} 1.65 & \cellcolor{pink} 1.88 &  \cellcolor{SkyBlue} 1.85 & \cellcolor{SkyBlue} 1.46 & \cellcolor{SkyBlue} 1.32 & \cellcolor{SkyBlue} 5.44 & \cellcolor{SkyBlue} 2.52 & \cellcolor{SkyBlue} 7.51 & \cellcolor{SkyBlue} 2.28 \\
        \midrule
        \textbf{DKT-SMoE} & \cellcolor{pink} 1.17 & \cellcolor{pink} 1.56 & \cellcolor{pink} 1.47 & \cellcolor{pink} 1.63 & \cellcolor{SkyBlue} 1.46 & \cellcolor{SkyBlue} 1.21 & \cellcolor{SkyBlue} 1.12 & \cellcolor{SkyBlue} 4.38 & \cellcolor{SkyBlue} 2.04 & \cellcolor{SkyBlue} 7.13 &\cellcolor{SkyBlue} 1.99 \\ 
        \bottomrule
    \end{tabular}
\end{table*}

\begin{table}
    \centering
    \scriptsize
    \caption{Impacts of the kernel sizes of Adapter layers and the effectiveness of SMoE.}
    \vspace{-0.3cm}
    \setlength{\tabcolsep}{2mm}{
    \begin{tabular}{c|cc|cc|cc}
    \toprule
    \multirow{2}{*}[-1pt]{Setting}  & \multicolumn{2}{c|}{\textbf{KITTI 2015}}  &  \multicolumn{2}{c|}{\textbf{Middlebury}} & \multicolumn{2}{c}{\textbf{ETH3D}} \\
    & \textbf{EPE} & \textbf{D1\_All}  & \textbf{EPE} & \textbf{Bad 2.0}  & \textbf{EPE}  & \textbf{Bad 1.0} \\
    \midrule
    Kernel Size = 3 & \textcolor{blue}{\textbf{0.62}} & \textcolor{blue}{\textbf{1.54}}  & \textcolor{blue}{\textbf{0.74}} & \textcolor{blue}{\textbf{4.26}} & 0.16 & 0.69\\
    Kernel Size = 5 & 0.66 & 1.62 & 0.77 & 4.61 & \textbf{0.15}  & 0.68 \\
    Kernel Size = 7 & 0.68 & 1.74 & 0.81 & 4.66 & 0.17 & 0.75 \\
    Kernel Size = 9 & 0.64 & 1.62 & 0.76 & 4.41 & \textbf{0.15} &\textcolor{blue}{\textbf{0.66}} \\
    \midrule
    SMoE & \textbf{0.60} & \textbf{1.51} & \textbf{0.71} & \textbf{4.12} & \textbf{0.15} & \textbf{0.63} \\
    \bottomrule
    \end{tabular}}
    \vspace{-0.2cm}
    \label{supp:cnn}
\end{table}

\begin{table}[t]
\scriptsize
\caption{Zero shot performance comparisons on different baselines.}
\vspace{-0.2cm}
\setlength{\tabcolsep}{1.2mm}{
    \begin{tabular}{c|c|cccc}
    \toprule
    \multirow{2}{*}[-1pt]{\textbf{Models}} & {VFM} & \multirow{1}{*}[-1pt]{\textbf{KIT 2012}}  &  \multirow{1}{*}[-1pt]{\textbf{KIT 2015}}  &  \multirow{1}{*}[-1pt]{\textbf{Middle}} &  \multirow{1}{*}[-1pt]{\textbf{ETH3D}}\\
    & Capacity & \textbf{Bad 3.0} & \textbf{Bad 3.0} & \textbf{Bad 2.0}  & \textbf{Bad 1.0}\\
    \midrule   
    PSMNet~\cite{chang2018pyramid} & -  & 6.0 & 6.3 & 15.8 & 10.2 \\
    PSMNet-SMoE & ViT-Base~\cite{yang2024depth_v2}  & {\textbf{4.1}} & {\textbf{4.9}} & {\textbf{8.3}} & {\textbf{5.4}} \\
    \midrule
    CFNet~\cite{shen2021cfnet} & - & 4.7 & 5.8 & 15.3  & 5.8 \\
    CFNet-SMoE & ViT-base~\cite{yang2024depth_v2} & \textbf{4.0} & \textbf{4.8} & \textbf{7.4} & \textbf{3.2}\\
    \midrule
    IGEV~\cite{xu2023iterative} & - & 5.1 & 5.6 & 7.1 & 3.6 \\
    IGEV-SMoE & ViT-base~\cite{yang2024depth_v2} & \textbf{4.1} & \textbf{4.7} & \textbf{6.8} & \textbf{2.0} \\
    \bottomrule
    \end{tabular}}
\label{supp:tab_baselines}
\vspace{-0.2cm}
\end{table}

\noindent\textbf{Compatibility of Different Baselines.}
Our SMoE design enhances the generalization performance of different baselines. The zero-shot results of all baselines are consistently improved within our SMoE framework, as detailed in Tab.~\ref{supp:tab_baselines}. 
For example, the Bad-error rate of PSMNet~\cite{chang2018pyramid} on each dataset decreases by 32\%, 22\%, 37\%, 47\%, and 48\%, respectively. A significant improvement is achieved in Middlebury since there is much abundant semantic information for reasoning. A similar improvement can be observed even when the robust CFNet~\cite{shen2021cfnet} is utilized as the baseline,  with the Bad-error rate being
decreased by 15\%, 17\%, \%, 52\%, and 45\%, respectively.
Furthermore, we observe that integrating our SMoE into the IGEV~\cite{xu2023iterative} baseline results in a significant improvement in cross-domain performance.
Notably, our approach does not require any specialized losses or additional modules to enhance domain generalization performance, making it broadly applicable to most learning-based stereo frameworks.

\noindent \textbf{The Effectiveness of Kernel Sizes of Adapter Layers.}
As introduced in Section 3, the designed CNN adapters with different receptive fields incorporate local geometry priors of input samples into the ViT block. 
From Table~\ref{supp:cnn}, we present a comparative analysis of the cross-dataset performance of the proposed SMoEStereo, utilizing individual Adapter experts with varying kernel sizes. Notably, different datasets demonstrate distinct optimal Adapter experts.
Our SMoE method adeptly identifies the most suitable local feature extractor for each input, achieving superior results compared to employing a single fixed Adapter. It is important to note that the SMoE approach introduces only negligible additional activated parameters, underscoring that the performance improvements are attributed to the proposed MoE learning scheme rather than scaling up the model.
In summary, these findings comprehensively validate the effectiveness of our SMoE design.

\begin{figure}[t]
    \centering
\includegraphics[width=1\linewidth]{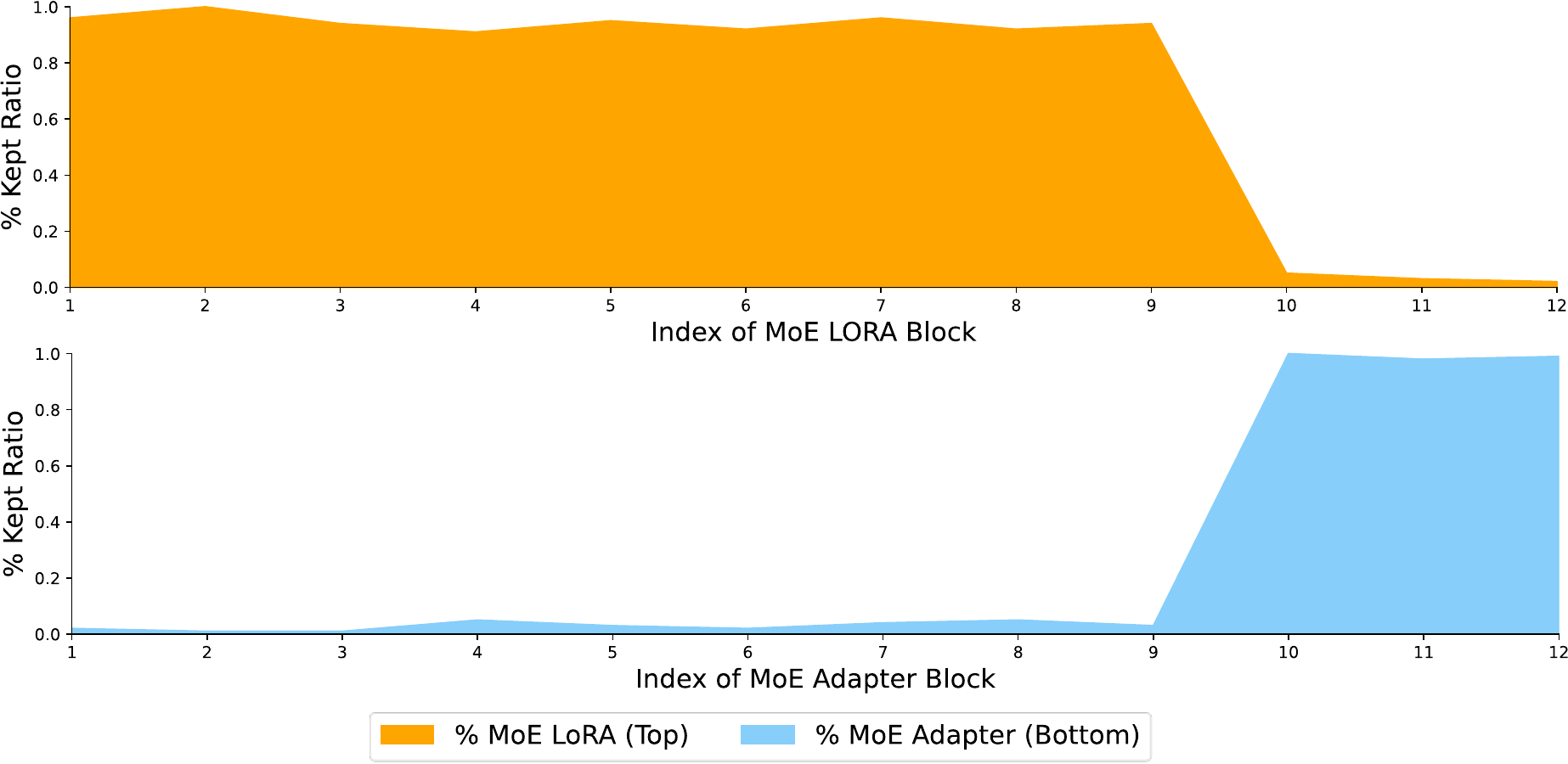}
\vspace{-0.6cm}
    \caption{Computational cost throughout the network. The kept/activated ratio of MoE LoRA modules (top) and MoE Adapter modules (bottom) throughout the backbone are reported.}
    \label{supp:index}
    \vspace{-0.2cm}
\end{figure}

\begin{figure*}[t]
    \centering
    \includegraphics[width=1\linewidth]{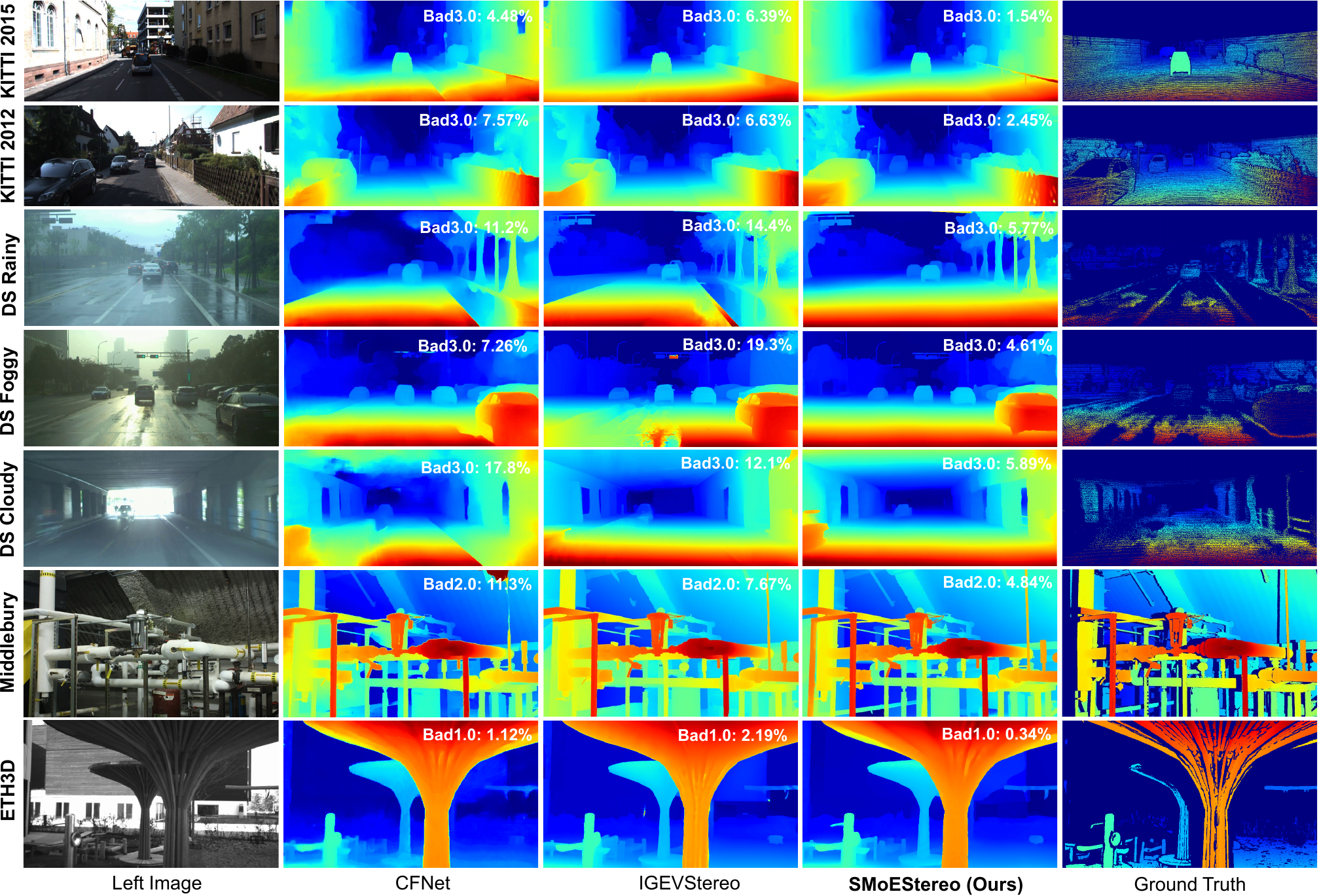}
    \vspace{-0.6cm}
    \caption{Zero-shot performance of CFNet~\cite{shen2021cfnet}, IGEVStereo~\cite{xu2023iterative}, and SMoEStereo on diverse scenes. Note that, all models are only trained on the SceneFlow dataset. In the middle row, DS denotes the DrivingStereo dataset.}
    \label{supp:visual}
    \vspace{-0.3cm}
\end{figure*}
\noindent \textbf{Computational Saving throughout the Network.}
SMoEStereo leverages computational redundancy to enhance the efficiency of VFMs. We collect usage policies on MoE LoRA and MoE Adapter selection predicted by our method across four real-world datasets, illustrating the distribution of computational cost (i.e., percentage of MoE LoRA/Adapter retained) throughout the backbone. As shown in Fig.~\ref{supp:index}, SMoEStereo strategically allocates more computation to MoE LoRA layers in the earlier stages of the network while reserving MoE Adapter layers for the latter stages.
This allocation suggests an optimization where MoE LoRA layers, which require less computational effort, handle initial processing, and MoE Adapter layers, functioning as decoders, manage more complex tasks in later stages of the ViT layers. This approach not only balances the computational load but also ensures efficient processing and superior performance across different scenarios, highlighting the robustness and adaptability of SMoEStereo.

\begin{figure*}[t]
    \centering
    \includegraphics[width=1\linewidth]{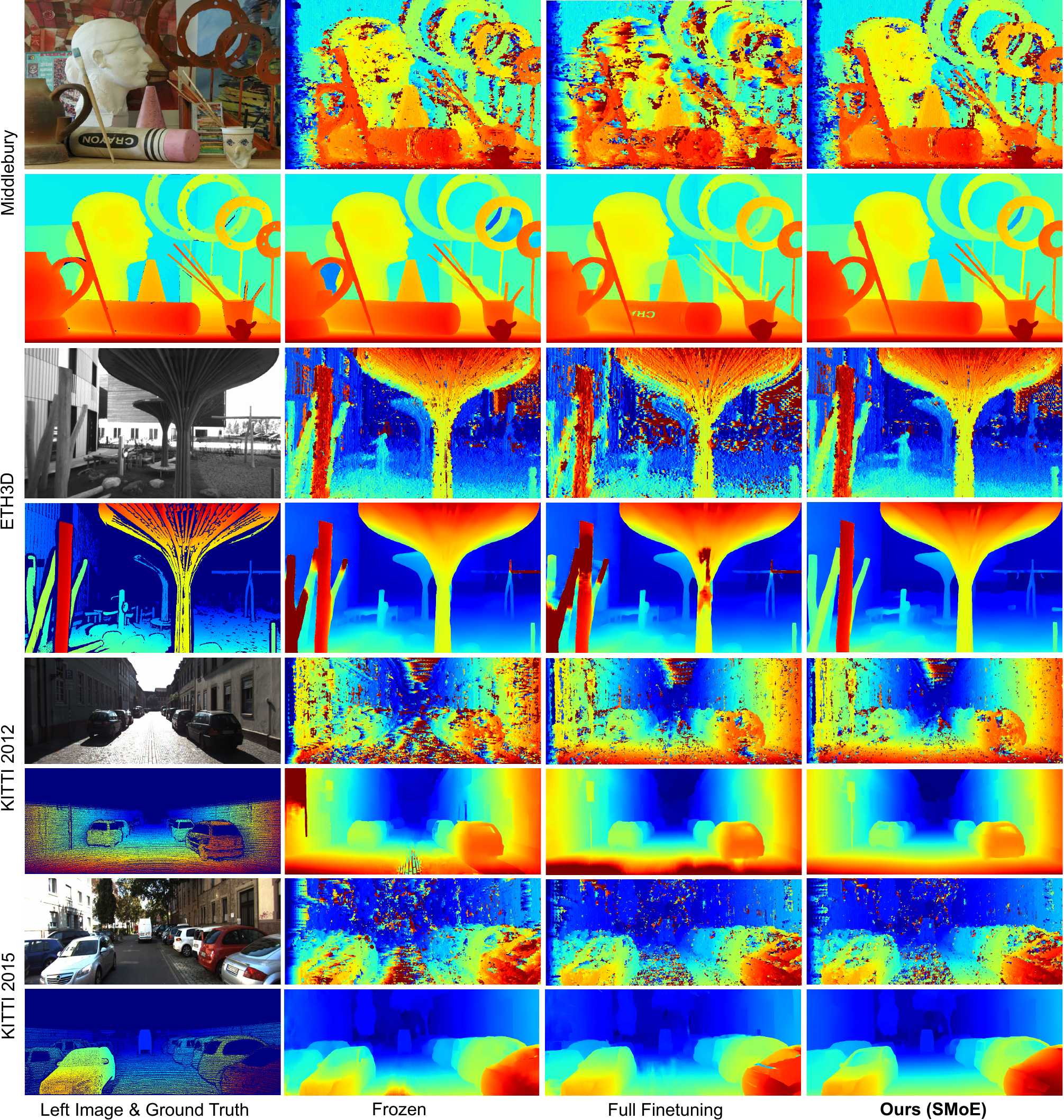}
    \vspace{-0.6cm}
    \caption{Zero-shot visual comparison of Frozen, Full-Finetuning, and our SMoE on diverse scenes. Note that, all models are only trained on the SceneFlow dataset.
    The rows (1 \& 3 \& 5 \& 7) indicate Winner-Take-All (WTA) disparity from the feature correlation (1/4 scale) achieved by the dot products among left and right features before the subsequent cost aggregation network.
    WTA disparity, when enhanced with our SMoE-equipped pre-trained VFM, exhibits significantly less noise compared to other vanilla finetuning methods for VFMs.}
    \label{supp:wta}
    \vspace{-0.3cm}
\end{figure*}

\noindent \textbf{The Effectiveness of MoE Balance Loss.}
The router network often assigns disproportionately large weights to a few experts~\cite{shazeer2017outrageously}, leading to overfitting issues. To counteract this, we introduce an MoE balance loss component to ensure equal importance among all experts, preventing the model from getting trapped in local optima.
In this subsection, we evaluate the impacts of the proposed $\mathcal{L}_{blc}$. Table~\ref{supp:loss} presents the cross-dataset evaluation performance with different values of the $\mathcal{L}_{blc}$ loss weight $\lambda_{1}$ in Eq. (11), where $\lambda_{1}$ = 0 represents no MoE loss applied. The application of $\mathcal{L}_{blc}$ effectively mitigates router overfitting and consistently enhances model generalizability. The model achieves optimal generalizability with $\lambda_{1}$ set at 1.
These findings underscore the critical importance of balanced expert contributions within the MoE framework. The $\mathcal{L}_{blc}$ component ensures that the learned representations are more diverse and generalizable across different datasets. This is especially crucial in practical applications where data distribution can vary significantly. The optimal performance observed at $\lambda_{1}=1$ suggests that balancing expert utilization and regularization is vital.

\label{sec:ablation}
\begin{table}[t]
    \centering
    \scriptsize
    \caption{Effectiveness of the proposed loss components.}
    \vspace{-0.3cm}
    \setlength{\tabcolsep}{2.7mm}{
    \begin{tabular}{c|cc|cc|cc}
    \toprule
    \multirow{2}{*}[-1pt]{$\lambda$}  & \multicolumn{2}{c|}{\textbf{KITTI 2015}}  &  \multicolumn{2}{c|}{\textbf{Middlebury}} & \multicolumn{2}{c}{\textbf{ETH3D}} \\
    & \textbf{EPE} & \textbf{D1\_All}  & \textbf{EPE} & \textbf{Bad 2.0}  & \textbf{EPE}  & \textbf{Bad 1.0} \\
    \midrule
    $\lambda=0$ & 0.69 & 1.68 & 0.79 & 4.74 & 0.18 & 0.82 \\
     $\lambda=0.1$ & 0.65 & 1.60 & 0.75 & 4.42 & 0.17 & 0.74 \\
      $\lambda=0.5$ & \textbf{0.60} & \textbf{1.48} & \textcolor{blue}{\textbf{0.74}}  & \textcolor{blue}{\textbf{4.17}} & 0.15 & 0.67 \\
     $\lambda=1$ & \textbf{0.60} & \textcolor{blue}{\textbf{1.51}} & \textbf{0.71} & \textbf{4.12} & \textbf{0.15} & \textbf{0.63} \\
     $\lambda=5$ & 0.63 & 1.57 & 0.74 & 4.25 & \textbf{0.16} & \textcolor{blue}{\textbf{0.71}} \\
    \bottomrule
    \end{tabular}}
    \label{supp:loss}
    \vspace{-0.3cm}
\end{table}

\noindent \textbf{Gains from More Synthetic Data.}
In the main paper, we trained our model solely on the SceneFlow dataset~\cite{mayer2016large}. 
This section investigates how increased data capacity affects SMoEStereo's performance with additional training samples. 
We consider two additional synthetic datasets: Virtual KITTI2~\cite{cabon2020virtual}, a synthetic outdoor driving dataset with 20K samples, and CREStereo~\cite{li2022practical}, a synthetic dataset with diverse delicate structures.
Table~\ref{supp:data} demonstrates that performance on KITTI 2012/2015 and ETH3D consistently improves with more datasets used for training. However, performance on Middlebury~\cite{middlebury2014} deteriorates when VKITTI2 samples are included. 
We argue that this is due to distinct domain shifts in the additional data, weakening generalization on Middlebury. Conversely, using the CREStereo dataset for training improves generalization on Middlebury.
In summary, these results suggest that SMoEStereo's generalization ability is enhanced with increased training data capacity.

\begin{table}[t]
    \centering
    \scriptsize
    \caption{Gains from data capacity. SF, VK2, and CRE denote the SceneFlow, Virtual KITTI2, and CREStereo datasets.}
    \scriptsize
    \vspace{-0.3cm}
    \setlength{\tabcolsep}{1.5mm}{
    \begin{tabular}{ccc|c|c|c|c}
    \toprule
    \multirow{2}{*}[-1pt]{SF} & \multirow{2}{*}[-1pt]{VK2}  &\multirow{2}{*}[-1pt]{CRE} & \textbf{KITTI 2012}   & \textbf{KITTI 2015}  &  \textbf{Middlebury} & \textbf{ETH3D} \\
    & & & \textbf{D1\_All}  &  \textbf{D1\_All} & \textbf{Bad 2.0} & \textbf{Bad 1.0} \\
    \midrule
    $\surd$ & - & - & 4.22 & 4.86 &     `7.05 & 2.10 \\
    $\surd$ & $\surd$ & - & {3.21} & {3.90} & 7.43 & 2.19 \\
    $\surd$ & $\surd$ & $\surd$ & \textbf{3.15} & \textbf{3.78} & \textbf{6.79} & \textbf{1.90} \\
    \bottomrule
    \end{tabular}}
    \label{supp:data}
    \vspace{-0.3cm}
\end{table}

\section{Zero-shot Performance on Diverse Scenarios.} 
We illustrate the predicted disparity maps across various scenarios in Fig.~\ref{supp:visual} for a qualitative comparison. Compared to previous state-of-the-art methods~\cite{shen2021cfnet,xu2023iterative}, our method shows exceptional generalization across diverse scenes, including outdoor, indoor, and challenging weather conditions. Additionally, our approach produces significantly less noisy disparity maps compared to vanilla finetuning methods for VFMs, showcasing enhanced robustness and effectiveness.
This improvement highlights our method's ability to maintain finer details and generate smoother results, particularly in zero-shot settings, as illustrated in Figure~\ref{supp:wta}. 
The robust features produced by our SMoE fully demonstrate the superiority of our approach in enhancing the overall zero-shot performance of stereo matching outcomes, ensuring fine detail preservation and smoothness even in complex and varied environments.

{
    \small
    \bibliographystyle{ieeenat_fullname}
   \bibliography{main}
}